%% file: main.tex
\documentclass{article} % For LaTeX2e
\usepackage{collas2022_conference,times}
\input{math_commands.tex}

\usepackage{hyperref}
\hypersetup{
    colorlinks=true,
    linkcolor=red,
    filecolor=magenta,      
    urlcolor=blue,
    citecolor=purple,
    pdftitle={Practical tradeoffs between memory, compute, and performance in learned optimizers},
    pdfpagemode=FullScreen,
    }

\usepackage{booktabs}       % professional-quality tables
\usepackage{amsfonts}       % blackboard math symbols
\usepackage{nicefrac}       % compact symbols for 1/2, etc.
\usepackage{microtype}      % microtypography
\usepackage{xcolor}         % colors
\usepackage{hyperref}

\usepackage{wrapfig}
\usepackage{overpic}
\usepackage[utf8]{inputenc} % allow utf-8 input
\usepackage[T1]{fontenc}    % use 8-bit T1 fonts
\usepackage{hyperref}       % hyperlinks
\usepackage{booktabs}       % professional-quality tables
\usepackage{amsfonts}       % blackboard math symbols
\usepackage{nicefrac}       % compact symbols for 1/2, etc.
\usepackage{microtype}  
\usepackage{amsmath}
\usepackage{varwidth}
\usepackage{lscape} 
\usepackage{url}            % simple URL typesetting

\definecolor{darkgreen}{rgb}{0,0.6,0}
\definecolor{darkred}{rgb}{0.7,0.0,0}
\definecolor{darkblue}{rgb}{0,0.0,0.6}
\definecolor{magenta}{rgb}{0.8,0.1,0.8}
\definecolor{darksomething}{rgb}{0,0.4,0.6}

\newcommand{\sref}[1]{\S\ref{#1}}

\usepackage{color}
\definecolor{deepblue}{rgb}{0,0,0.5}
\definecolor{deepred}{rgb}{0.6,0,0}
\definecolor{deepgreen}{rgb}{0,0.5,0}

\title{Practical tradeoffs between memory, compute, and performance in learned optimizers}

\collasfinalcopy

\author{%
  Luke Metz \\
  Google Research, Brain Team\\
  \texttt{lmetz@google.com} 
  \And
  C. Daniel Freeman \\
  Google Research, Brain Team\\
  \texttt{cdfreeman@google.com} \\
  \And
  James Harrison \\
  Google Research, Brain Team \\
  \texttt{jamesharrison@google.com} \\
  \And
  Niru Maheswaranathan \\
  Meta \thanks{Work done while at Google Research}\\
  \texttt{niru@hey.com} \\
  \And
  Jascha Sohl-Dickstein \\
  Google Research, Brain Team\\
  \texttt{jaschasd@google.com}\\
}

\begin{document}

\maketitle

\begin{abstract}
Optimization plays a costly and crucial role in developing machine learning systems. 
In learned optimizers, the few hyperparameters of commonly used hand-designed optimizers, e.g. Adam or SGD, are replaced with flexible parametric functions. 
The parameters of these functions are then optimized so that the resulting learned optimizer minimizes a target loss on a chosen class of models.
Learned optimizers can both reduce the number of required training steps and improve the final test loss.
However, they can be expensive to train, and once trained can be expensive to use due to computational and memory overhead for the optimizer itself. 
In this work, we identify and quantify the design features governing the memory, compute, and performance trade-offs for many learned and hand-designed optimizers. 
We further leverage our analysis to construct a learned optimizer that is both faster and more memory efficient than previous work.
Our model and training code are open source\footnote{\url{https://github.com/google/learned_optimization}}.
\end{abstract}

\section{Introduction}
Despite the huge computational costs associated with training large neural models, the set of optimization algorithms used to train them has largely been restricted to simple update functions mapping from gradients to parameter updates (e.g. stochastic gradient descent \citep{robbins1951stochastic}, Adam \citep{kingma2014adam}, or RMSProp \citep{tieleman2012lecture}).
These algorithms typically depend on a small number of hand-designed features and parameters.
However, the last decade in machine learning research has repeatedly seen small, hand-designed models outperformed by parameterized models (such as neural networks) trained to purpose on large amounts of data \citep{lecun2015deep}.
Thus, a promising direction to improve training performance and reduce costs is to replace hand-designed optimizers with more expressive {\em learned optimizers}, trained on problems similar to those encountered in practice. 

Learned optimizers specify parameter update rules using a flexible parametric form and learn the parameters of this function from a ``dataset'' of optimization tasks---a procedure typically referred to as meta-training or meta-learning \citep{andrychowicz2016learning, finn2017model, hochreiter2001learning}.
Learned optimizers represent a path towards improved optimizer performance, and possess the ability to target different objectives (e.g. test loss \citep{metz2019understanding}, or robustness \citep{metz2019using}), as well as the ability to leverage new features useful for optimization.
Despite being an active area of  research~\citep{andrychowicz2016learning, wichrowska2017learned, chen2020training, metz2020using, metz2021training, almeida2021generalizable, zheng2022symbolic}, they are not yet commonly used in practice.
Several challenges have limited the widespread application of learned optimizers: they are typically difficult to meta-train on a task family of interest, they can require significant memory and compute overhead when applied, and they often generalize less well to novel tasks than hand-designed optimizers.

In this work, we aim to precisely study the limitations of learned optimizers, and address these limitations via a novel learned optimizer architecture. 
In particular, we explore and quantify the tradeoffs in terms of memory, compute, and generalization across a variety of optimizers, including hand-designed optimizers~\citep{bottou2010large, tieleman2012lecture, kingma2014adam},
learned hyper-parameter controllers~\citep{daniel2016learning, hansen2016using, xu2017reinforcement, xu2019learning, almeida2021generalizable},
and neural network based learned optimizers~\citep{andrychowicz2016learning, wichrowska2017learned, metz2020tasks}, with the goal of understanding how choices in optimizer design affect performance and usability.
Our core contributions are:
\begin{enumerate}
    \item We present a thorough empirical characterization of the trade-offs inherent in different learned optimizer architectures and features, and a comparison of these learned optimizer architectures against their well-tuned hand-designed counterparts.
    \item We develop a new per-parameter learned optimizer architecture, on the Pareto frontier with regards to performance, computational cost, and memory usage among existing learned, and hand designed optimizers. 
    \item We provide an open source implementation written in JAX \citep{jax2018github} to enable future research and reliable benchmarking\footnote{\url{http://github.com/google/learned_optimization}}.
\end{enumerate}

\section{Optimizers}
In this section we review and formalize the class of optimizers that are commonly used in training neural networks. We then define meta-learned optimizers, and highlight differences with standard optimization approaches. We describe several examples of both common, standard neural network optimizers as well as classes of learned optimizers, all of which are investigated in this paper. 

\subsection{Gradient Based Optimizers}

Most first-order optimizers\footnote{Optimizers using only gradient information and not higher order derivatives.} used to train neural networks can be viewed as functions mapping from a history of gradients to parameter updates.
We will assume the optimizer acts on an underlying model with parameters $\phi \in \Phi$, while maintaining an internal optimizer state $s \in S$.
The parameters may be, for example, neural network weights, whereas the optimizer state includes quantities such as 
the accumulated momentum values in momentum-accelerated optimizers \citep{polyak1964some, nesterov1983method}.
The optimizer acts by ingesting gradients $g$
(which arise from a specified loss function and a dataset)
and outputting updated parameters $\phi'$.

More precisely, we define an optimizer as a pair of functions.
The first, which we call the \textit{Update} function, computes new parameter values $\phi'$ and state $s'$ from stochastic gradients, the current parameter value, and the current optimizer state. The second, which we refer to as the \textit{Init} function, initializes the optimizer state. Both functions have hyperparameters $\theta \in \Theta$, such as the learning rate or the initial value of accumulators. Thus, we write the optimizer as:
\begin{align}
    \text{Optimizer} :: (\text{Init}, \text{Update}) \\
    \text{Init} :: & (\phi; \theta) \rightarrow s \\
    \quad \text{Update} :: & (\phi, g, s; \theta) \rightarrow (s', \,\, \phi')
\end{align}

Optimizers can benefit from problem information beyond stochastic gradients, parameter values, and losses.
For instance, methods that utilize line searches~\citep{le2011optimization}, validation loss~\citep{xu2019learning, metz2020tasks}, or the structure of the underlying computation graph~\citep{martens2015optimizing} all rely on additional information.
However, the present work is restricted to optimizers which minimize training loss by mini-batch stochastic gradient descent.

%%% begin rewrite.
\textbf{First-order hand-designed optimizers:} \label{sec:hand designed archs}
Hand-designed optimizers typically have a simple form, and a small number of hyperparameters ($\theta$), which are tuned by random search \citep{bergstra2012random}, Bayesian optimization \citep{snoek2012practical}, or other low-dimensional black-box optimization techniques \citep{bergstra2011algorithms, golovin2017google, optuna_2019}.
They mostly have low overhead in terms of compute and memory usage.
For instance, Adam \citep{kingma2014adam} has two accumulators, and SGD has none\footnote{Some hand-designed methods, such as Shampoo \citep{gupta2018shampoo, anil2020second}, involve considerable compute overhead, but can make more progress per update step.}.

In this work, we experiment with four kinds of hand-designed optimizers: SGD~\citep{robbins1951stochastic, bottou2010large}, SGDM (SGD with momentum) \citep{polyak1964some}, Adam~\citep{kingma2014adam}, and Nesterov accelerated Adam ~\citep{dozat2016incorporating} with AdamW~\citep{loshchilov2017decoupled} style weight decay (NAdamW).
For SGD, SGDM, and Adam, we search over learning rates every half order of magnitude between $10^{-7}$ and $1$. For SGM we set momentum to $0.9$ and for Adam we set $\beta_1=0.9$, $\beta_2=0.999$, and $\epsilon=10^{-8}$.
For NAdamW we use random search with many more hyperparameter configurations per task (1000) and a much larger search space over $8$ hyperparameters controlling: first and second momentum time scales, weight decays, and learning rate schedules.
Past work has shown this to be a powerful search space~\citep{metz2020using} and, in our work, this dramatically outperforms learning rate search.
See Appendix~\ref{app:nadamw_form} for more details.
Many other hand-designed optimizer architectures have been proposed \citep{ruder2016overview, zeiler2012adadelta, reddi2018adaptive, you2019large, liu2019variance}, but their practical benefits are small in most situations \citep{schmidt2020descending}.

\textbf{Factorized optimizers:}
In some settings, having even one additional copy of parameters to use for accumulators is too costly.
Recent methods such as AdaFactor~\citep{shazeer2018adafactor} and SM3~\citep{anil2019memory} factorize the weights and accumulate statistics using a sub-linear amount of memory with respect to parameters.
This style of accumulator has not been explored in the context of learned optimizers, but we will show this provides an effective way to improve performance without meaningfully increasing memory overhead (\sref{sec:arch_sweep}). 

\subsection{Meta-Learned Optimizers}
The meta-learning problem for optimizers consists of tuning the hyperparameters $\theta$ of a class of parameterized optimizers with respect to some loss function\footnote{Common choices of loss function for the meta-optimization problem include the average training loss across inner optimizer iterations, the average validation loss, as well as the terminal train/validation loss.}.
How is this different from the hyperparameter tuning discussed in the last subsection?
While there is no formal difference between the hyperparameter selection problem and training learned optimizers, the learned optimizers we consider in this subsection universally include a black-box component with a (comparatively) large number of parameters (in our case, always parameterized by a neural network).
This large number of parameters limits the effectiveness of traditional hyperparameter tuning methods such as random search, and so we focus on local optimization methods (including first-order gradient-based methods as well as zeroth-order methods) which are able to perform better in high dimensional optimization.
Below, we outline several types of learned optimizer. 

\textbf{Hyperparameter controllers:} \label{sec:hparam_controller}
Optimizing the hyperparameters of a hand-designed optimizer over a broad set of tasks may limit the performance within each specific task. These hand-designed optimizers can be augmented with a meta-learned controller, often parameterized as a neural network, that modulates the hyperparameters of the optimizer over the course of training to yield better performance in each particular problem~\citep{daniel2016learning, hansen2016using, xu2017reinforcement, xu2019learning, almeida2021generalizable}. 
This controller takes in summary statistics (e.g. gradient norms, loss values), and can either globally assign identical hyperparameters to all layers, or operate per-layer.
One benefit of hyperparameter controllers is that their per-parameter compute overhead is small, as the majority of the computation only needs to be performed once per tensor, or per network rather than scaling with the number of parameters.

We introduce a novel hyperparameter controller architecture which we refer to as \textbf{nn\_adam}. This architecture consists of an  LSTM-based~\citep{hochreiter1997long}  hyperparameter controller, operating on features derived from each tensor independently, and outputting Adam hyperparameters consisting of a per-tensor learning rate, $\beta_1$, $\beta_2$, and $\epsilon$. 
For features, this model uses normalized values derived from the first moment of gradients, the second moment, and the tensor shape.
See Appendix~\ref{app:nn_adam_arch} for details.

\textbf{Per-parameter learned optimizers:}
Per-parameter learned optimizers \citep{andrychowicz2016learning} learn a function, often parameterized by a neural network, which is applied to each parameter independently, though sometimes with normalization performed across parameters in a tensor~\citep{metz2019understanding}.

\textbf{Multi-level approaches:}
In an effort to add additional capacity to a learned optimizer while retaining good computational complexity with respect to number of parameters, hierarchical models have been proposed \citep{wichrowska2017learned, metz2020tasks}. 
These models leverage up to three levels of hierarchy: a \textit{global controller}, which sends and receives activations from a \textit{per-layer (or per-tensor) controller}, which finally sends and receives activations to a \textit{per-parameter optimizer}.

\textbf{New per-parameter learned optimizer:} \label{sec:lolv4_arch}
Finally, we introduce a new learned optimizer architecture (which we call \textbf{small\_fc\_lopt}) that combines architectural features of per-parameter and factorized optimizers, and outperforms both. 
This architecture will be directly motivated by the trade-offs among compute, memory, performance, and generalization shown in \sref{sec:experiments}. 
Our learned optimizer incorporates an extremely {\em tiny}, per-parameter, MLP-based learned optimizer similar to that used in \citet{metz2019understanding}.
This 197 parameter MLP takes as input 39 input features with 4 per-parameter accumulators (3 momenta at different meta-learned timescales, and 1 gradient second moment accumulator), and 3 AdaFactor accumulators also at 3 different meta-learned timescales. These features are passed into a 1 hidden layer, 4 hidden unit MLP.
See Appendix~\ref{app:lolv4_details} for additional details.

\section{Training and Meta-Training}

In the previous section we specified possible architectures for standard optimizers (with a small number of hyperparameters) as well as learned optimizers.
Both learned and hand designed optimizers are iteratively applied to some parameterized model, paired with a loss function and (possibly) a dataset. We refer to this collection as a \textit{task}.
We use the loss obtained by an optimizer on these tasks to select hyper-parameters (in the case of hand designed optimizers), and to optimize the learned optimizer weights.
In this section, we discuss the tasks used, the measurement of performance by which we can compare optimizers (meta-loss), and discuss how the weights of the learned optimizers are computed (which we refer to as meta-optimization).

\subsection{Tasks} \label{sec:tasks}
Throughout this paper, the tasks of interest are neural network training problems.
Each task is specified via three quantities.
The first is the underlying model architecture and the initial parameter values (or a procedure for initializing the model parameters).
The second is a function to generate batches of data, and the third is a loss function.
While a more abstract definition of a task could cover more general optimization problems, we aim to address neural network training as a setting and believe generalizations are (in most cases) straightforward.
In this work we consider solely supervised learning. 
We also consider only a single function to generate a batch of data, though this could easily be extended to multiple functions corresponding to, for example, train and validation loss. As discussed in the next subsection, we focus solely on training loss for simplicity. 

We primarily consider two tasks in this paper: A 2 hidden layer MLP with 128 hidden units and ReLU activations on Fashion MNIST \citep{xiao2017/online}, and a 3 layer convolutional network on CIFAR-10 \citep{krizhevsky2009cifar}.
See Appendix~\ref{app:inner_tasks} for more details and implementations. In Section \ref{sec:outer_generalize}, to assess generalization, we additionally evaluate optimizers meta-trained on these two tasks on three additional problems.

The tradeoffs inherent in optimizer design are task dependent (see \sref{sec:compute_density}), and the per-parameter compute and memory requirements of the optimizer must be balanced against the per-parameter compute and memory requirements of the task.
These latter requirements are a function of parameter sharing, sparsity in parameter use, model architecture, and minibatch size (compute overhead per parameter can be made arbitrarily small by increasing the minibatch size). 
The two problems we consider have different compute and memory requirements and were therefore chosen as reasonable baseline tasks providing insight into optimizer performance at different points in the space of possible tasks. Moreover, these (relatively small) tasks were chosen to enable the large-scale evaluation and comparisons done in this paper.

\subsection{Meta-Loss and Meta-Optimization}
To evaluate an optimizer, we apply our optimizer for 2,000 iterations and evaluate the average loss obtained over the course of training.
In this work, we exclusively focus on training loss performance as opposed to validation loss. This is to decouple optimization performance tradeoffs from the implicit regularization effects of learned optimizers shown in \citet{metz2019understanding, metz2020tasks}.

We train optimizers targeting the two tasks described above by randomly sampling from a fixed search space for hand-designed optimizers, and Persistent Evolution Strategies (PES) \citep{pmlr-v139-vicol21a} to train the learned optimizers.
See Appendix~\ref{app:course_details} for details.
To minimize confounds, we focus on the scenario where meta-train matches meta-test (i.e. the tasks presented during training and testing are the same), and examine the overhead and performance tradeoffs inherent in a learned optimizer meta-trained to optimize a single task. 

\section{Exploring tradeoffs across optimizer families} \label{sec:experiments}

In this section we experimentally explore tradeoffs when designing learned optimizers. In \sref{sec:course}, we show memory and time trade-offs for various hand-designed and learned optimizers.
In \sref{sec:arch_sweep}, we focus on per-parameter learned optimizers and explore the impact of both feature choice and size of the learned optimizer. 
In \sref{sec:compute_density}, we discuss computational costs of running learned optimizers as a function of task features (such as the number of network weights). In \sref{sec:unroll_length}, we tie all these evaluations together and show wall-clock time performance for the different tasks.
In \sref{sec:outer_generalize}, we explore meta-generalization---applying optimizers to a task different from those in which they were meta-trained.

\subsection{Compute, memory, performance tradeoffs for learned and hand-designed optimizers} \label{sec:course}
We characterize the trade-offs between performance, memory overhead, and compute overhead for both hand-designed and learned optimizers. 
The optimizers examined here consist of the hand-designed optimizers (\textbf{SGD}, \textbf{Adam}, and \textbf{NAdamW}), the MLP optimizer from \citet{metz2019understanding} (\textbf{fc\_lopt}), the hierarchical optimizer from \citet{metz2020tasks},
%[rnn, fc, lopt]
(\textbf{rnn\_fc\_lopt}), 
a hyperparameter controller described in \sref{sec:hparam_controller} (\textbf{nn\_adam}), and the per-parameter optimizer proposed in \sref{sec:lolv4_arch} (\textbf{small\_fc\_lopt}).

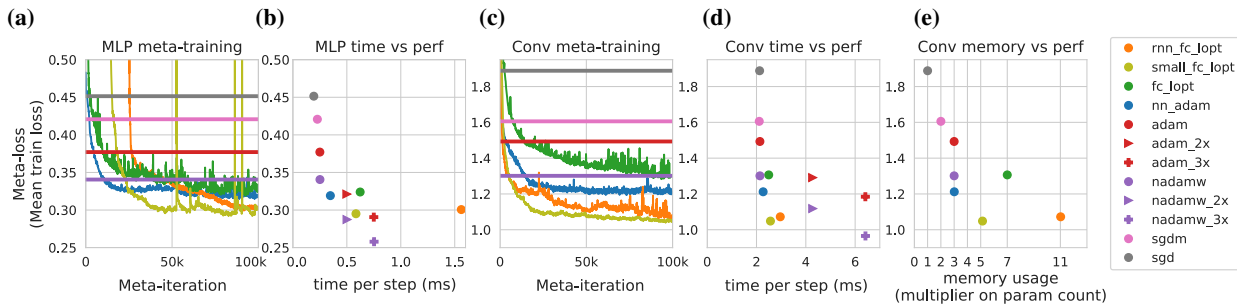
\begin{figure}
    \centering
\begin{overpic}[width=1.0\textwidth]{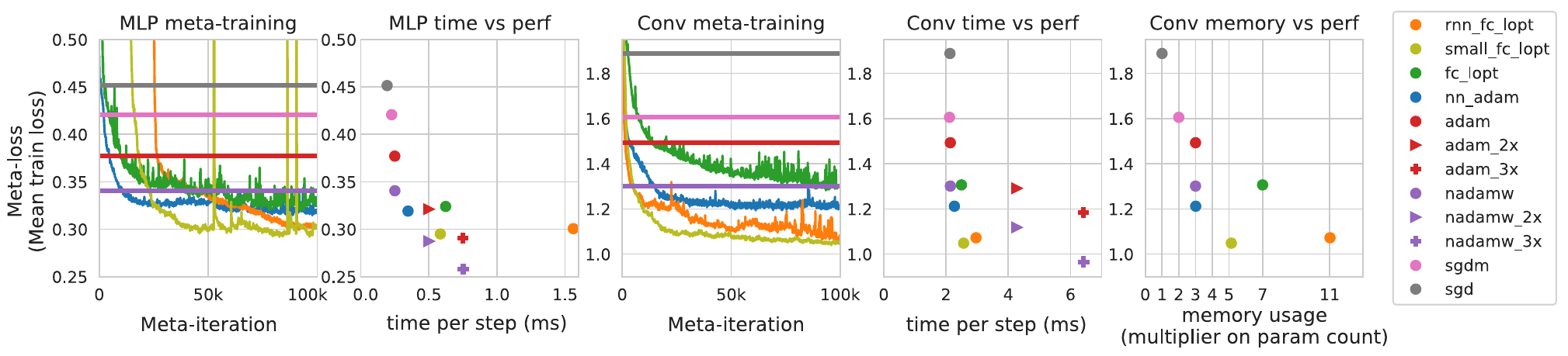}
\put (0,23.0) {\textbf{\small(a)}}
\put (20,23.0) {\textbf{\small(b)}}
\put (38,23.0) {\textbf{\small(c)}}
\put (56,23.0) {\textbf{\small(d)}}
\put (73,23.0) {\textbf{\small(e)}}
\end{overpic}
\vspace{-2em}
\caption{
    \textbf{Optimizer overhead depends on problem specification} We show two different tasks with large, and small, overheads.
    \textbf{(a,c)} Meta-learning curves targeting training an MLP on Fashion MNIST, and a ConvNet on CIFAR-10, respectively. 
    \textbf{(b,d)} Per-iteration run-time vs achieved meta-loss for different optimizers for the MLP and ConvNet respectively. For hand designed optimizers (horizontal bars) we show the best performing hyperparameters.
    Additionally computed is adam\_2x, adam\_3x, nadamw\_2x, and nadamw\_3x which show Adam and NAdamW run for 2x and 3x as many iterations.
    \textbf{(e)} Memory vs. best meta-loss for each optimizer. The new learned optimizer we introduce, small\_fc\_lopt, and our Adam controller baseline nn\_adam, lie on the Pareto frontier with respect to both memory and compute time among the optimizers and tasks tested.
    \label{fig:course}
    }
\end{figure}

Meta-training curves are shown in Figure~\ref{fig:course}ac.
We additionally show final performance of the fully trained learned optimizer as a function of compute time per step (Figure~\ref{fig:course}bd) and with respect to memory usage (Figure~\ref{fig:course}e).
We find learned optimizers can achieve lower meta-loss than baselines, but at the cost of more compute time and memory usage.
For the MLP task, the cost of the learned optimizer far outstrips the cost of a hand-designed optimizer (taking > 5x more time in the case of rnn\_fc\_lopt). For the CIFAR-10 ConvNet however, the compute overhead is small relative to overall compute, due to the much larger per-parameter cost for computing gradients for a ConvNet.

\subsection{Design choices for the MLP learned optimizer} \label{sec:arch_sweep}

\begin{figure}
    \centering
\begin{overpic}[width=1.0\textwidth]{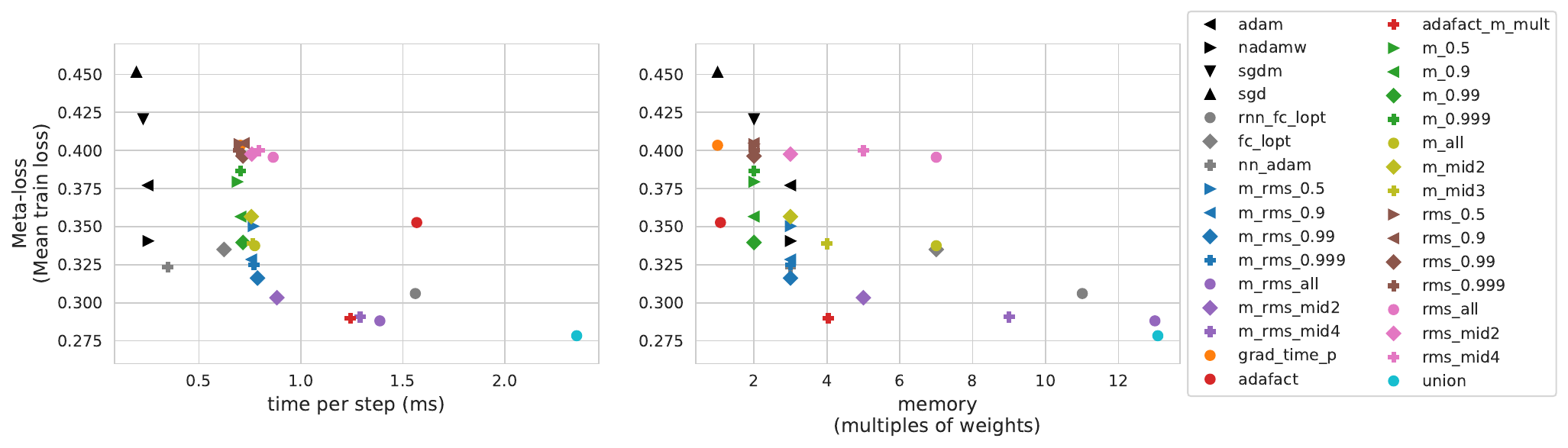}
\put (0,28) {\textbf{\small(a)}}
\put (40,28) {\textbf{\small(b)}}
\end{overpic}
    \vspace{-2em}
    \caption{\textbf{Trade-offs in performance, compute, and memory overhead for different learned optimizer architectures.} \textbf{(a)} Performance vs. compute and \textbf{(b)} performance vs. memory is shown for many choices of input features for an MLP-based learned optimizer, as well as for baselines consisting of both hand-designed (black) and previously published learned optimizers (gray). 
    The best raw performance is achieved by the learned optimizer configuration with the largest memory and compute overhead.
    \label{fig:mlp_features}
}
\end{figure}

To guide learned optimizer design, we explore the memory, time, and performance trade-offs associated with different choices of input features for an MLP learned optimizer.
The dominant source of memory overhead is the inclusion of additional per-parameter accumulators. 
We explore two kinds of per-parameter accumulators that optimizers use---the exponential moving average of the gradient's first moment, and second moment as used in momentum and RMSProp \citep{tieleman2012lecture} respectively.
Unlike existing optimizers, the learned optimizers we explore accumulate these statistics over multiple timescales%
\footnote{This is similar to what is done in the AggMo \citep{lucas2018aggregated} hand-designed optimizer.}.
In addition to these, we also explore preconditioning features based on AdaFactor \citep{shazeer2018adafactor} which use sub-linear memory in parameter count.

We plot performance vs. compute cost, and performance vs. memory, in Figure~\ref{fig:mlp_features}. 
We plot baselines with hyperparameters found via random grid search (in black) (adam, sgd, sgdm, nadamw), and baseline learned optimizers (in \textcolor{gray}{gray}) (fc\_lopt \citep{metz2019understanding}, rnn\_fc\_lopt~\citep{metz2020tasks}, nn\_adam (\sref{sec:hparam_controller}).
All other conditions consist of differently parameterized learned optimizers with different input feature. Each configuration is trained with PES \citep{pmlr-v139-vicol21a} for 100k meta-training iterations.
We test optimizers using only a single momentum accumulator with different decays (in \textcolor{green}{green}),
multiple momentum accumulators (in \textcolor{yellow}{yellow}),
a single second moment accumulator (in \textcolor{brown}{brown}),
multiple second moment accumulators (in \textcolor{pink}{pink}),
two accumulators with the same decay for first and second moments (in \textcolor{blue}{blue}),
multiple decay first and second moments (in \textcolor{purple}{purple}),
using AdaFactor features with and without additional momentum accumulators (in \textcolor{red}{red}),
using only gradient features (in \textcolor{orange}{orange}),
and finally the union of all features (in \textcolor{gray}{gray}).
See Appendix~\ref{app:mlp_features_details} for more experimental details.

We find the general trend that providing more features to a learned optimizer leads to better performance.
However, including more accumulators increases the computational and memory overhead of using these optimizers. 
AdaFactor features by themselves (\textbf{adafact}) use very little memory, but do not perform well. Combining a small number of momentum features with AdaFactor features (\textbf{adafact\_m\_mult}) recovers the performance of using second moment accumulators, without the need for second moment accumulators.

\begin{figure}
    \centering
\begin{overpic}[width=0.8\textwidth]{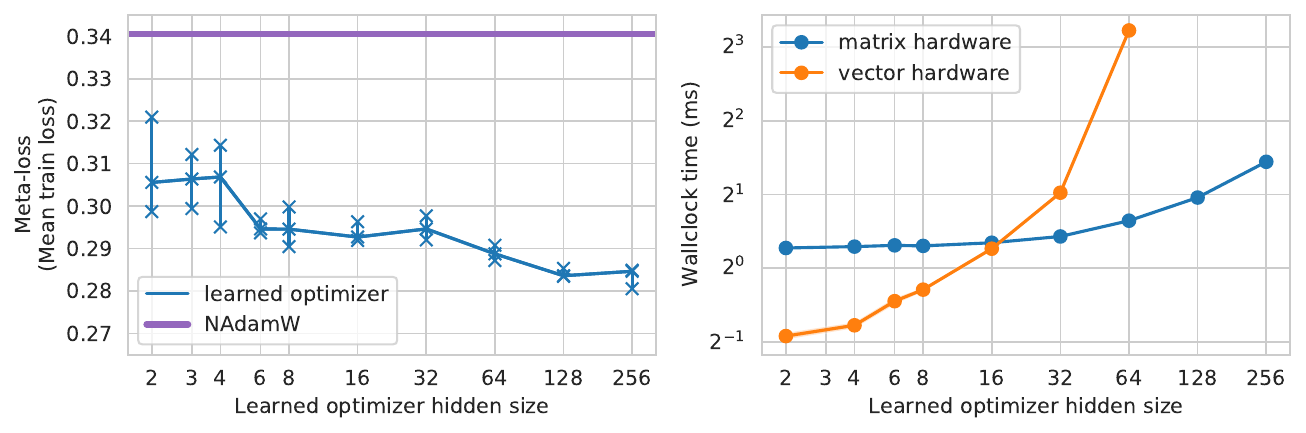}
\put (3,32.0) {\textbf{\small(a)}}
\put (53,32.0) {\textbf{\small(b)}}
\end{overpic}
\vspace{-1em}
\caption{\textbf{Trade-offs in performance and compute overhead as a function of learned optimizer size.}
A grid search is performed over the hidden state size of an MLP-based learned optimizer, for a fixed set of input features (based on small\_fc\_lpot, \sref{sec:lolv4_arch}). 
    \textbf{(a)} Performance vs. hidden state size for the MLP-based optimizer, with a baseline of the best hand-designed optimizer (nadamw, \sref{sec:hand designed archs}).
    Each 
    ``$\times$''
    denotes a random seed.
    As hidden size increases, performance improves, and variance in performance decreases.
    \textbf{(b)} Compute time measured on a TPUv2 as a function of performance achieved for different hidden sizes for two different low-level implementations -- one using on-accelerator matrix multiplication hardware, and the other using on-accelerator vector hardware.
    For matrix hardware, we find roughly constant performance up to 64 hidden units, whereas with vector hardware, we see speedups as hidden size is reduced all the way to two units.
    \label{fig:sizes}}
\end{figure}
Finally, we explore varying the hidden size of the MLP (Figure~\ref{fig:sizes}).
Using the same features as in small\_fc\_lpot (\sref{sec:lolv4_arch}), we sweep the hidden size of the MLP from 2 to 256 units.
For each width we perform a small hyperparameter search over meta-learning rate selecting between $3\cdot10^{-5}$,$10^{-4}$, $3\cdot10^{-4}$ and take the best performing learning rate for each width.
Surprisingly, an extremely narrow MLP is sufficient to outperform the best hand-designed baseline (NAdamW).
Increasing width boosts performance, but performance improvements diminish.

The relationship between learned optimizer width and compute overhead
depends heavily on implementation details.
TPU and other accelerator hardware often have specialized matrix multiplication units that operate on fixed dimensional matrices (e.g. TPUv2 has 128x128 systolic arrays \citep{norrie2020google}).
For a naive implementation of the learned optimizer using matrix multiplication kernels on TPUv2, there are no significant speedups from shrinking the optimizer width below approximately 64 units. These matrix are too small and thus don't fully utilize the matrix units.
If matrix multiplication is instead expanded explicitly in terms of element-wise operations, then continued speedups can be achieved even for optimizer hidden state vectors of two units, achieving a nearly 2x speedup over the use of matrix multiplication primitives as these element wise computations use a different subset of the accelerator hardware for the same computation.
Profiling suggests that even greater speedups should be possible using custom kernels (as are frequently written for hand-designed optimizers).
In a sense, these tiny learned optimizers, with matrix multiplications expanded, blur the line between hand designed and learned optimizers as both implementations are a handful of element wise floating point operations.

\subsection{Overhead of learned optimizers on different tasks} \label{sec:compute_density}

\begin{figure}
    \centering
\begin{overpic}[width=1.0\textwidth]{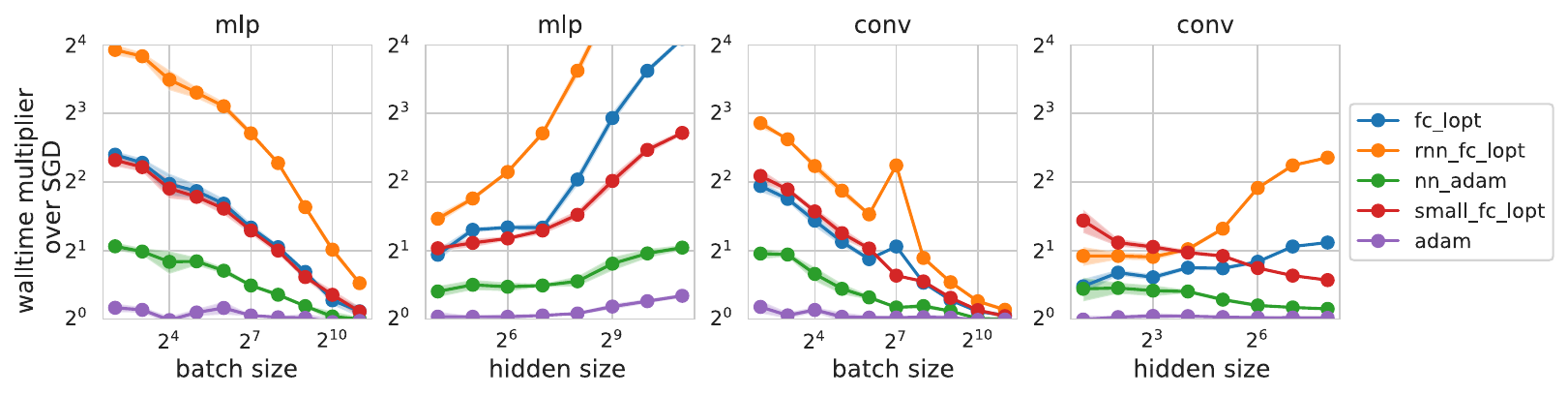}
\put (4,24.0) {\textbf{\small(a)}}
\put (26,24.0) {\textbf{\small(b)}}
\put (48,24.0) {\textbf{\small(c)}}
\put (67,24.0) {\textbf{\small(d)}}
\end{overpic} %
    \vspace{-1em}
    \caption{
    \textbf{Optimizer compute overhead shrinks with increasing batch size, and has a model-specific interaction with model size.} 
    Plot shows overhead of different optimizers relative to training with SGD, as a function of width and batch size for both an MLP and ConvNet. 
    Shaded regions denote standard deviation over 5 measurements.
    \label{fig:compute_density}
    }
\end{figure}
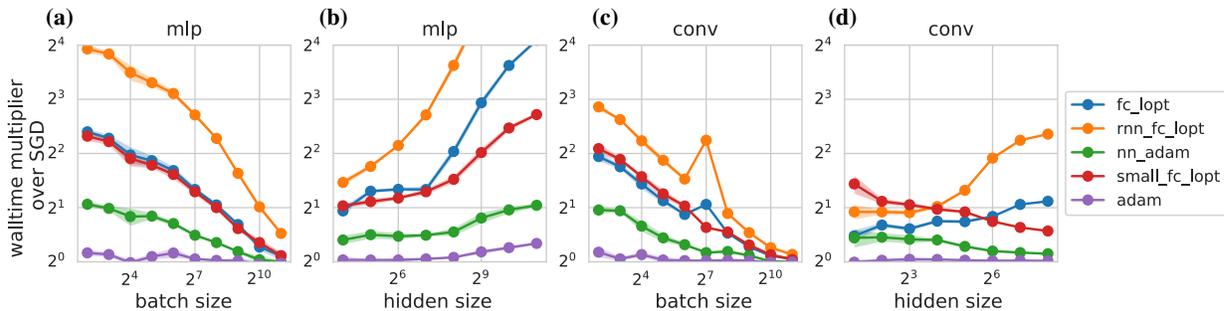

To explore the dependence on task identity,
we measure the relative overhead of training with a learned optimizer compared to SGD for different widths and batch sizes of the Fashion MNIST MLP and CIFAR-10 ConvNet.
Results are shown in Figure~\ref{fig:compute_density}.
We find that, in all cases, increasing batch size lowers the overhead, as the cost to compute gradients increases but the cost of applying the optimizer remains constant.
In the case of the ConvNets, increased the hidden size (channel count) of the target problem lowers the overhead for small\_fc\_lopt and nn\_adam, but increases for fc\_lopt and rnn\_fc\_lopt. 
For MLP target problems, increasing the target MLP's hidden size increases overhead for all optimizers including Adam.
The asymptotic scaling of this behavior is due to both the computational complexity, and memory bandwidth, of the underlying hardware.

Next, we explore overheads for some common, large scale models.
Table \ref{table:tasks} show results for ResNets and Transformers, trained on a single TPUv2 chip.
Distributed training would allow us to split the optimizer computation across devices and thus achieve even lower optimizer overhead.
See Appendix \ref{app:large_timing} for further details.
\begin{table}[]
\caption{\textbf{On realistic tasks, the overhead of a learned optimizer is typically small.} We show parameter count, time per step for SGD, and overhead of the small\_fc\_lopt learned optimizer for four tasks: two different sized ResNets, with different batch sizes (BS), as well as two ``small'' Transformers~\citep{vaswani2017attention} with different word sequence lengths (L) and batch sizes (BS). All numbers are medians over 10 timings. Standard error is under the reported number of digits. \label{table:tasks}}
\centering
\small
\begin{tabular}{|l|l|l|l|l|}
\hline
Task                    & Params & SGD step time (ms) & LOpt step time (ms) & LOpt multipler \\ \hline
ResNet18(BS=128)        & 11.7M  & 159 & 180 & 1.13           \\ \hline
ResNet50(BS=32)         & 25.6M  & 99.5 & 137.9 & 1.39           \\ \hline
Transformer(L=256,BS=16) & 43.1M  & 91.2 &  132.4 & 1.45             \\ \hline
Transformer(L=512,BS=2) & 43.1M  & 29.7  &  70.9 & 2.39              \\ \hline
\end{tabular}
\end{table}

\subsection{Performance with respect to wall clock time} \label{sec:unroll_length}

\begin{figure}
    \centering
\begin{overpic}[width=1.0\textwidth]{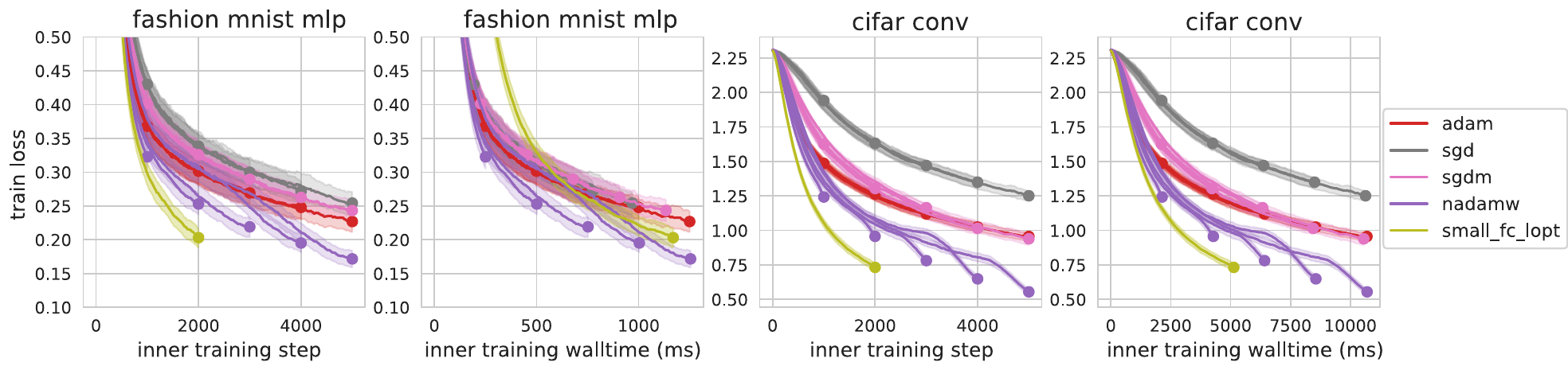}
 \put (5,22.0) {\textbf{\small(a)}}
 \put (26,22.0) {\textbf{\small(b)}}
 \put (48,22.0) {\textbf{\small(c)}}
 \put (69,22.0) {\textbf{\small(d)}}
\end{overpic}
    \vspace{-1em}
    \caption{
    Inner-training curves for \textbf{(a,b)} the Fashion MNIST MLP task, and \textbf{(c,d)} the CIFAR-10 ConvNet task. 
    We show training loss with respect to iteration and wall clock time. 
    For each baseline, we show runs where the optimizer was meta-trained to achieve the best loss for varying inner-training lengths. 
    Solid circles denote the final training performance after the number of training steps targeted for that baseline. For example, the yellow curve in (a) stops at 2000 steps because this small\_fc\_opt was meta-optimized for performance at 2000 inner training steps.
    Shaded region show standard deviation across 10 seeds.
    \label{fig:walltime}}
\end{figure}
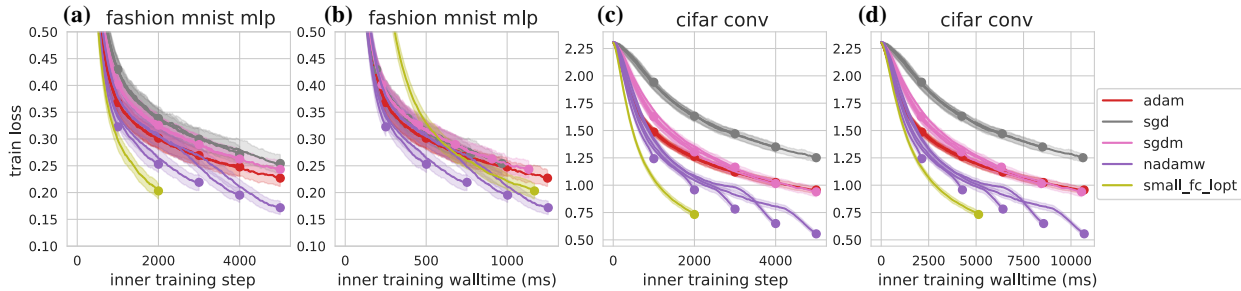

In the previous sections, we measured the performance achieved by optimizers, and the computational overhead required to achieve that performance.
In practice, one often cares most about the total wall clock time required to reach a given performance. 
Further, the optimal meta-parameters $\theta$ change depending on the length of inner-training. 
To quantify the achievable performance as a function of wall clock time, 
we compare training trajectories for both our learned and hand-designed optimizers. 
We apply a learned optimizer meta-trained for length 2k unrolls, and optimize hyperparameters of hand-designed optimizers to perform well on 1k, 2k, 3k, 4k, and 5k length unrolls. 
In Figure~\ref{fig:walltime} we show the resulting performance for both the Fashion MNIST MLP task and CIFAR-10 ConvNet tasks.
Our learned optimizer is always faster with respect to step count. 
With respect to wall-clock time, on the CIFAR-10 ConvNet task we also see faster training, while for the Fashion MNIST MLP task, where the relative overhead of the learned optimizer is large, NAdamW performs best.

\subsection{Meta-generalization: optimizer performance on holdout tasks} \label{sec:outer_generalize}

\begin{figure}
    \centering
\begin{overpic}[width=1.0\textwidth]{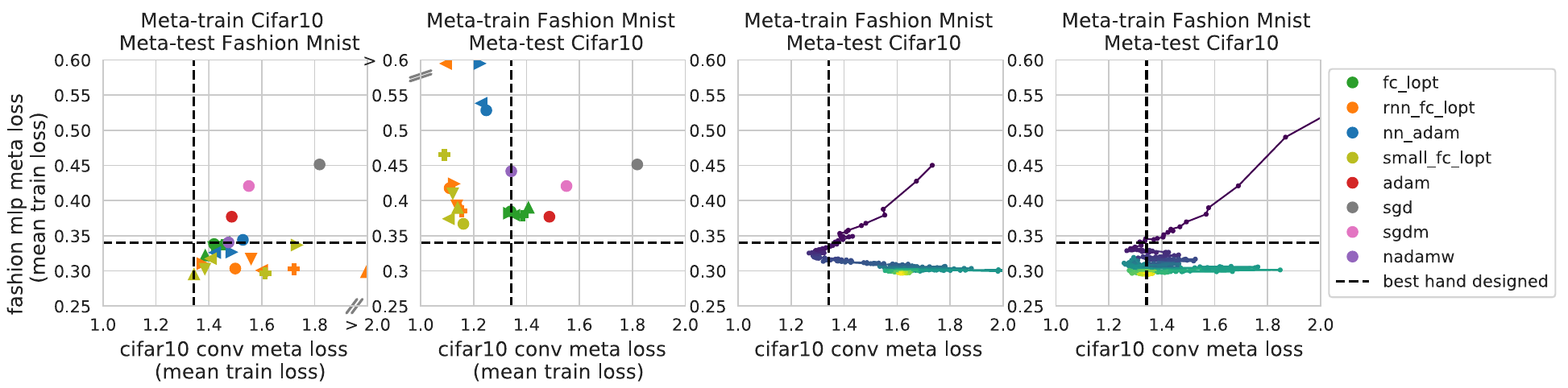}
\put (6,24.5) {\textbf{\small(a)}}
\put (27,24.5) {\textbf{\small(b)}}
\put (47,24.5) {\textbf{\small(c)}}
\put (70,24.5) {\textbf{\small(d)}}
\end{overpic}
    \vspace{-1em}
    \caption{
    \textbf{Meta-generalization and meta-overfitting of learned and hand-designed optimizers.}
    \textbf{(a)} Optimizers are meta-trained (/hyperparameter-tuned) on the Fashion MNIST MLP task, and tested on the CIFAR-10 ConvNet task. 
    \textbf{(b)} Optimizers are meta-trained on the CIFAR-10 ConvNet task, and tested on the Fashion MNIST MLP task. 
    Each marker represents a different optimizer. 
    Dashed lines denote best tuned performance across all hand-designed optimizers on the task. 
    \textbf{(c,d)} Meta-training trajectory for the ``small\_fc\_lopt'' optimizer on the Fashion MNIST MLP task. The $y$-axis shows the meta-loss (performance after training the target task with the learned optimizer) on the Fashion MNIST MLP task, while the $x$-axis shows the meta-loss on the CIFAR-10 ConvNet task. 
    Purple is early in meta-training, yellow is at the end of meta-training. 
    Each pane shows a different random seed.
    Early in meta-training the learned optimizer generalizes, outperforming hand-designed optimizers on both the meta-training and meta-testing task. 
    As meta-training continues, the learned optimizer meta-overfits, doing better on the Fashion MNIST MLP task but worse on the CIFAR-10 ConvNet task.
    \label{fig:meta_generalize}}
\end{figure}
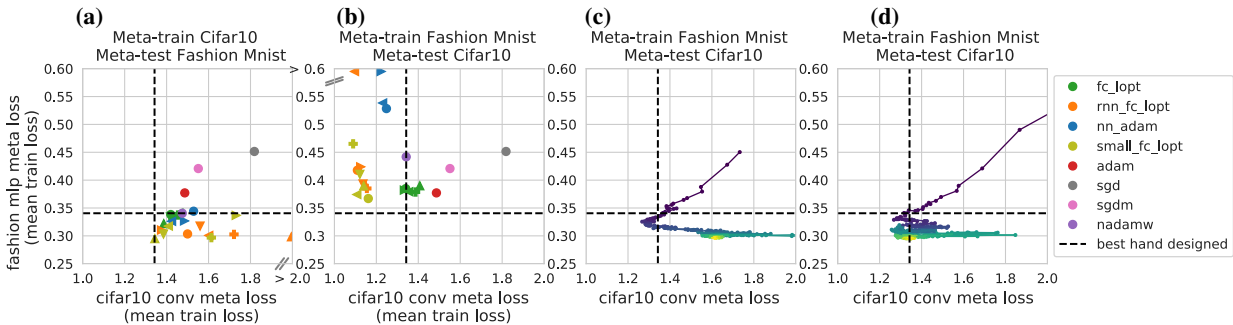

One final trade-off is the interaction between optimizer design choice and generalization performance. 
Generalization performance, in this context, refers to the ability of the optimizer to perform well at training a novel task, different from the task distribution used to meta-train the optimizer.
To quantify this, we measure the performance of diverse optimizers meta-trained on one task, but then used as the optimizer for a novel task. 

We meta-train each of the different learned optimizers on either the Fashion MNIST MLP
or the CIFAR-10 ConvNet tasks. Over the course of meta-training we evaluate performance on 5 tasks:
\begin{enumerate}
    \item The Fashion Mnist MLP described in \ref{sec:tasks}.
    \item The CIFAR-10 ConvNet described in \ref{sec:tasks} using 16x16 images for computational reasons.
    \item A three hidden layer  MLP trained on 16x16 imagenet.
    \item A CIFAR-10 auto-encoder trained with mean squared error loss.
    \item An LSTM~\citep{hochreiter1997long} language model trained on byte level LM1B\citep{DBLP:journals/corr/ChelbaMSGBK13}.
\end{enumerate}
See Appendix \ref{app:more_generalization} for more details and implementations.

First, we train a number of learned optimizers of different types and select the checkpoint which performs best on the meta-training task, and evaluate their performance on different held out tasks. We show transfer performance when meta-training and meta-testing on Fashion MNIST MLP and CIFAR-10 ConvNet in Figure~\ref{fig:meta_generalize}, and the remainder of the comparisons in Appendix~\ref{app:more_generalization}. While there is some correlation across learned optimizer architecture, in general we find poor meta-generalization.
Additionally, meta-generalization seems to depend more strongly on details of the meta-training process than it does on learned optimizer architectural choices. 
This variation poses challenges when reporting results due to the cost of meta-training.

To explore variance in the meta-training process, we plot performance over the course of meta-training on both the target-task, and the held out tasks.
We show these dynamics for two different learned optimizer seeds in Figure~\ref{fig:meta_generalize}cd.
In both cases, meta-training trajectories exhibit high variability. 
However, they also both show an initial phase of correlated performance improvement, culminating in better performance than the baselines for both the target and held out task, before the optimizer finally overfits to the target-task.

\begin{figure}
    \centering
\begin{overpic}[width=1.0\textwidth]{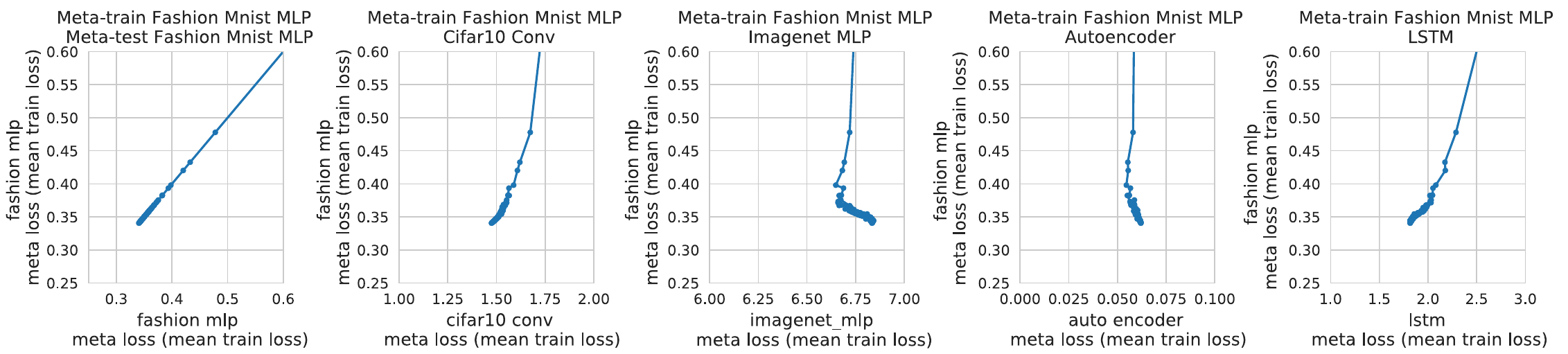}
\end{overpic}
    \vspace{-1em}
    \caption{We show meta-training vs meta-test performance after, for a given budget, finding hyper-parameter for the NAdamW optimizer on the meta-training task (in this case Fashion MNIST MLP) and evaluating the best found hyperparameter for the given budget on various held out tasks.
    Similar to learned optimizers, we find meta-overfitting. We see this when evaluating on the Imagenet MLP and the Autoencoder but not when transferring to CIFAR-10 ConvNet nor the LSTM. \label{app_fig:meta-gen-nadam}}
\end{figure}

This type of meta-overfitting is not unique to learned optimizers and happens even when trying to transfer hand designed optimizers from task to task. To show this, we simulate meta-training on the Fashion MNIST MLP by randomly sampling subsets of parameters from the original NAdamw search space for different budgets. We then find the best performance on the meta-training task and apply the best hyperparameters to the meta-test task. We show meta-train vs meta-test performance for different budgets in Figure~\ref{app_fig:meta-gen-nadam}. We see signs of meta-overfitting for some tasks, such as the ImageNet MLP and the MLP Autoencoder. For the others, CIFAR-10 ConvNet and LSTM, we continue to see correlation between meta-train and meta-test.

\section{Related Work}
Originally proposed in \citet{bengio1992optimization, runarsson2000evolution}, interest in learned optimizers has undergone a recent revival. 
Proposed learned optimizer architectures have included per-parameter RNNs \citep{andrychowicz2016learning}, hierarchical models enabling sharing of information across parameters \citep{wichrowska2017learned, metz2020tasks}, and a simplified architecture consisting just of an MLP \citep{metz2019understanding}. 
Optimizer meta-training techniques have included gradient descent \citep{maclaurin2015gradient, andrychowicz2016learning, wichrowska2017learned}, reinforcement learning \citep{li2016learning, li2017learning}, 
and more advanced training procedures \citep{ lv2017learning, maheswaranathan2019guided, metz2019understanding, chen2020training, pmlr-v139-vicol21a, metz2021training} leveraging both Evolution Strategies (ES), and gradients.
Learned optimizers have been targeted at applications including model robustness \citep{metz2019using}, chemistry \citep{learn2hop}, min-max optimization \citep{shen2021learning}, adversarial training \citep{xiong2020improved}, few-shot learning \citep{ravi2016optimization}, swarm optimization \citep{cao2019learning}, unsupervised learning \citep{metz2018learning}, black box optimization \citep{chen2016learning}, and MCMC sampling \citep{levy2017generalizing,wang2017meta,gong2018meta}.
Other work has analyzed learned optimizer behavior \citep{maheswaranathan2020reverse}.
\citet{Bello17} takes a different approach, and meta-learns symbolic rather than neural-network driven parameter update rules.

In an effort to understand computational costs, we look to Pareto frontiers of computation and memory vs performance. The concept of Pareto optimally was originally proposed in economics to understand how individuals can prosper with finite resources~\citep{newman1998new} and has since become a useful tool in computer science.
Studying trade-offs in this way is common in computer vision and natural language processing, where performance as a function of model size is often explored~\citep{simonyan2014very, he2016deep, vaswani2017attention}.
Building efficient frontiers of models has been a target for meta-learning as well \citep{tan2019efficientnet}.
In the scope of learned optimizers, \citet{metz2019understanding} explored training wall-clock efficiency, but on limited hardware (CPU) and with respect to a single target problem instance. \citet{wichrowska2017learned} showed that the relative overhead of computing updates with a learned optimizer shrinks as batch size is increases. \citet{zheng2022symbolic} propose a symbolic distillation meta-training step which converts neural network parameterized optimizers to a symbolic form resulting in both lower memory and compute costs. There has also been significant research exploring the trade-offs between different optimization techniques outside of deep learning, especially between stochastic, full batch, and between different second order methods. For example, Newtons method has led to a number of approximations -- e.g. diagonal approximations \citep{duchi2011adaptive}, block diagonal \citep{martens2015optimizing, gupta2018shampoo}, as well as a large family of quasi newton methods \citep{dennis1977quasi} (e.g. BFGS \citep{broyden1970convergence} and its low memory counterpart, L-BFGS \citep{liu1989limited}).

\section{Conclusion}
In this work, we characterized practical trade-offs involved in designing learned optimizers, including those between performance optimizing a target task, compute and memory overhead associated with the learned optimizer, training time, choice of target task, and generalization to new tasks. 
Using the lessons learned from our careful exploration,
we introduce an architecture that strikes a better balance between memory usage, compute, and performance. 
We then show that this learned optimizer architecture can be used to accelerate training on accelerator hardware.

The goal of this work was to provide a thorough investigation of the fundamental tradeoffs associated with learned optimizers. 
We view this paper as a first step toward the empirical characterization necessary for principled comparisons, but our experiments were limited in several ways. First, to control the number of covariates and perform the required experiments within a reasonable computation budget, we have limited ourselves to (primarily) two tasks, which themselves are simple compared with state-of-the-art neural network models. Moreover, we have provided only a limited investigation of meta-generalization and validation performance. In general, while this paper serves as a first step toward rigorous empirical comparison within this novel class of learned optimizers, further work is required to extend our results. 

In order to make it easier for future research to build on our work, and to include better grounded empirical comparisons, all optimizers, tasks, and training code are open sourced in
learned\_optimization\footnote{\url{https://github.com/google/learned_optimization}},
% redacted\_package\footnote{\url{https://github.com/google/learned_optimization}},
an open source library written in JAX for designing, training, and testing learned optimizers.

\section*{Acknowledgements}
We would like to thank Chip Huyen, Ben Poole, Amil Merchant, 
and Wenqing Zheng for their support and comments on this work, as well as the entire Brain team for providing a wonderful research environment. We would also like to thank the authors of the python scientific computing stack including Numpy \citep{van2011numpy}, and Matplotlib \citep{matplotlib}.
\clearpage

\bibliography{references}
\bibliographystyle{plainnat}

\appendix

\section{small\_fc\_lopt architectural details} \label{app:lolv4_details}
We describe the full details of our proposed learned optimizer. The source code for this optimizer can be found in \url{https://github.com/google/learned_optimization/blob/aa15091066aa5b3f45e6b7f4bee1c41fb7d467a0/learned_optimization/learned_optimizers/adafac_mlp_lopt.py}.

Our learned optimizer consists of features, concatenated then fed into an MLP. These features contain:
\begin{itemize}
    \item the parameter values
    \item the 3 momentum values ($m$)
    \item the second moment value ($v$)
    \item 3 values consisting of momenta normalized by rms gradient norm -- $m/\sqrt{v}$
    \item the $\left(\sqrt{\text{v}+\epsilon}\right)^{-1}$ value
    \item 3 AdaFactor normalized gradient values
    \item the tiled, AdaFactor row features (3 features)
    \item the tiled, AdaFactor column features (3 features)
    \item $1/\sqrt{\text{adafact feats}}$ of these previous 6 features
    \item 3 features consisting AdaFactor normalized momentum values
    \item  11 features formed by taking the current timestep, $t$, and computing $\text{tanh}(t/x)$ where $x \in \{ 1, 3, 10, 30, 100, 300, 1000, 3000, 10k, 30k, 100k \}$
\end{itemize}

All but the time features are normalized to have a second moment of 1 across the tensor and fed into a $4$ hidden unit, 2 hidden layer MLP with a ReLU activation function then projected to 2 hidden dimensions representing a magnitude, $m$, and a scalar direction $d$ to be combined to form a predicted step: $\Delta \phi = \lambda_1 d \text{exp}(\lambda_2 m)$ where $\lambda_1$ and $\lambda_2$ are constants set to 0.001 to keep initial step-sizes small.

In order to reduce computational overhead, this optimizer is dramatically smaller than past learned optimizers, containing only 197 meta-parameters.

\section{Tasks we meta-train on} \label{app:inner_tasks}
\subsection{Fashion MNIST MLP}
The MLP we use consists of 2 hidden layers with 128 hidden units and ReLU activations. It is trained on batches of Fashion MNIST re-scaled to lie between [0, 1] and on batch sizes of 128. We train with cross entropy loss.
Our network was built in Haiku \citep{haiku2020github} and the implementation can be found at \url{https://github.com/google/learned_optimization/blob/32c4f21ec238a12756afe70e3d699017ea938f5d/learned_optimization/tasks/fixed/image_mlp.py#L35}.

\subsection{CIFAR-10 ConvNet}
The ConvNet consists of 3 hidden layers with ReLU activations. All layers have kernel sizes of 3x3. The first layer has 32 units and stride 2. The following 2 layers have 64 hidden units and stride 1. All convolutions have same padding. We average over the spatial dimensions then linearly project to 10. We train on batches of 128 CIFAR-10 re-scaled between [0, 1] and use cross entropy loss. The implementation can be found at \url{https://github.com/google/learned_optimization/blob/ba2b56565fb507368652d2e4a12ab305a6d99ded/learned_optimization/tasks/fixed/conv.py#L96}.

\section{nn\_adam architecture} \label{app:nn_adam_arch}
The nn\_adam learned optimizer is a hyper parameter controller based learned optimizer. In addition to the description that follows, we provide an implementation at \url{https://github.com/google/learned_optimization/blob/ba2b56565fb507368652d2e4a12ab305a6d99ded/learned_optimization/learned_optimizers/nn_adam.py#L161}.
For each tensor of the target problem, we compute some set of features (see \ref{nnadam_features}), feed them into a 32 unit LSTM, and output 4 values -- log learning rate, beta 1, beta 2 (both parameterized as log(1 - beta)) and log epsilon. These hyperparameters are then fed into Adam where the update to each weight and accumulator follows the Adam update equations.

\subsection{per tensor features} \label{nnadam_features}
We use a same set of per-tensor features as used by the hierarchical learned optimizer in \citet{metz2020tasks}. For many features, we employ a simple transformation to obtain magnitudes of features. This transformation involves computing the log of the absolute value, clipping between -5 and 5, and rescaling by 0.5.

For each tensor, we use the following as a feature set:
\begin{itemize}
    \item Transformed mean momentum value
    \item Sign of mean momentum
    \item Transformed variance squared of momentum
    \item Transformed mean of the accumulator of second moment
    \item Sign of the mean of the accumulator of second moment
    \item Transformed mean parameter value
    \item Sign of mean parameter value
    \item Transformed variance squared of parameter value.
    \item Transformed mean gradient value
    \item Sign of mean gradient value
    \item Transformed variance squared gradient
    \item Transformed mean absolute value of gradient.
\end{itemize}

\section{Experiment details}
\subsection{Common} \label{app:meta-train_time}
For all experiments in this paper we meta-train with Persistent Evolutionary Strategies \citep{pmlr-v139-vicol21a} with a standard deviation of 0.01 and length 20 truncations with length 2k inner training steps.
The meta-objective we target is mean training loss (clipped at the initialization value -- $ln(10)$ in the classification problems).
We use 4 distributed workers (each using a single TPU accelerator chip) in an async batched fashion. We use a meta-batch size of 4 with each meta gradient being an average from each worker which is itself an average over 8 tasks.
For all models, we have an additional learner job (also a TPU chip) which averages meta-gradients and performs PES updates.
We use Adam as the meta-optimizer in all experiments with gradient clipping of 3.0 done on each value of the gradient independently.
Each training job has an additional 3 machines (15 for meta-generalization experiments), each with a single TPU chip to perform evaluations by training a task with the current meta-parameters.
The meta-training curves we report are from these machines.
All meta-training in this work took ~2-4 days per experiment.

\subsection{\sref{sec:course} details: Optimizer overhead vs optimizer type} \label{app:course_details}
For each task, and learned optimizer pair we train 3 random seeds for 5 learning rates: [1e-5, 3e-5, 1e-4, 3e-4, 1e-3].
We show the best performing optimizer. We do this as opposed to mean as the low dimensional hidden size of small\_fc\_lopt (4 hidden units) can result in unstable training.
We found using a simple learning rate schedule improves meta-training stability, but for a fair comparison to other optimizers we do not use any schedule here.

\subsection{\sref{sec:arch_sweep} details: Input features experiment details} \label{app:mlp_features_details}
For these experiments we use a fixed learning rate, set to $10^{-4}$ as this was found to be the best performing model for fc\_lopt from \sref{app:course_details}.
Given the amount of variations tried, we could not afford to search over learning rate for each configuration. For each configuration we compute 3 random seeds.

\textbf{grads\_time\_p}: Just using parameter, and gradient, and time step features.

\textbf{m\_0.1, m\_0.5, m\_0.9, m\_0.99, m\_0.999}: Using parameter value, gradient, momentum with the listed decay value, and time step features.

\textbf{m\_all}: Same as before but using multiple momentum values. In this case five values: 0.1, 0.5, 0.9, 0.99, 0.999.

\textbf{m\_mid2}: Same as before but with 2 momentum values 0.5 and 0.9.

\textbf{m\_mid3}: Same as before but with 3 momentum values 0.5, 0.9 and 0.99.

\textbf{rms\_0.1, rms\_0.5, rms\_0.9, rms\_0.99, rms\_0.999}: Using parameter value, gradient, the second moment accumulator with the listed decay value, 1 over the sqrt of this feature, and time step features.

\textbf{rms\_all}: Same as before but using multiple second moment accumulator values. In this case six values: 0.1, 0.5, 0.9, 0.99, 0.999, 0.9999.

\textbf{rms\_mid2}: Same as before but with 2 second moment values 0.9 and 0.99.

\textbf{rms\_mid4}: Same as before but with 3 second moment values 0.5, 0.9, 0.99, and 0.999.

\textbf{m\_rms\_0.1, m\_rms\_0.5, m\_rms\_0.9, m\_rms\_0.99, adams\_0.999}: Using parameter value, gradient, the second moment accumulator with the listed decay value, 1 over the sqrt of this feature, momentum with the listed decay, as well as the product of momentum and 1 over the square root of the second moment (similar the Adam update) and time step features.

\textbf{m\_rms\_all}: Same as before but using multiple second moment and momentum accumulator values. In this case six values: 0.1, 0.5, 0.9, 0.99, 0.999, 0.9999.

\textbf{m\_rms\_mid2}: Same as before but with 2 accumulator timescales: 0.9 and 0.99.

\textbf{m\_rms\_mid4}: Same as before but with 3 accumulator timescales: 0.5, 0.9, 0.99 and 0.999.

\textbf{adafact}: Using parameter values, 6 adafactor accumulator decays (0.1, 0.5, 0.9, 0.99, 0.999, 0.9999) which are fed to the learned optimizer in the form 3 multiplications: 1 over sqrt, 1 over sqrt multiplied by the gradient, and by tiling both of the low rank accumulators.

\textbf{adafact\_m\_mul}: Same as before, but with 3 adafactor accumulators and 3 momentum accumulators of decays (0.5, 0.9, 0.99). In addition to the previous features, we also include the multiplication of momentum value by the preconditioner from adafactor.

\textbf{union}:
The union of all features. This includes parameter value, gradient value, time features, all momentum and second moment accumulators (0.1, 0.5, 0.9, 0.99, 0.999, 0.9999), all features from adafactor computed with these same timescales, as well as multiplications of adafactor and momentum features.

\subsection{\sref{sec:compute_density} details: Large scale overhead timings} \label{app:large_timing}
We use the ResNet18, and ResNet50 implementations from Haiku \citep{haiku2020github} with the V2 flag set to true.

For transformers, we use vocab size of 256 to emulate byte level training, a hidden size of 768, 6 layers, and 12 self attention heads per layer. When applying dense layers we use a 4x widening factor.

% \clearpage

\section{Extended meta-generalization experiments}
\label{app:more_generalization}

\subsection{Experimental details}
Over the course of training a learned optimizer on a particular task, we monitor performance on a variety of held out tasks described here.

\textbf{Fashion Mnist MLP}: This is a 2 hidden layer, 128 unit MLP trained on fashion mnist. Source code can be found \url{https://github.com/google/learned_optimization/blob/32c4f21ec238a12756afe70e3d699017ea938f5d/learned_optimization/tasks/fixed/image_mlp.py#L35}.

\textbf{CIFAR-10 Convnet}: This is convnet with 3 hidden layers trained on 16x16 CIFAR-10. It contains 3 hidden layers starting with a 32 channels stride 2, and followed by two 64 channel, stride 1 convolutions. Average pooling is then performed before linearly mapping to the number of output channels. An implementation can be found at \url{https://github.com/google/learned_optimization/blob/78f25e8f1e9c6236a1f559b7b0b36859c59542d2/learned_optimization/tasks/fixed/conv.py#L86}

\textbf{Imagenet MLP}: This is an MLP operating on 16x16 resized imagenet images. The network has 3 hidden layers, of size 256. An implementation can be found at \url{https://github.com/google/learned_optimization/blob/aa15091066aa5b3f45e6b7f4bee1c41fb7d467a0/learned_optimization/tasks/fixed/image_mlp.py#L94}.

\textbf{Auto Encoder}: This is an auto encoder trained on CIFAR-10 with mean squared error. The network consists 3 hidden layers with sizes 128, 32, 128. A full implementation can be found in \url{https://github.com/google/learned_optimization/blob/aa15091066aa5b3f45e6b7f4bee1c41fb7d467a0/learned_optimization/tasks/fixed/image_mlp_ae.py#L101}.

\textbf{LSTM language modeling}: This is a language model trained on \citep{DBLP:journals/corr/ChelbaMSGBK13}. The language is tokenized as bytes, and sliced into length 32 sequences. The model consists of embedding the tokens with a 64 dimensional lookup table, followed by a size 128 LSTM tasked to predict the next token. See \url{https://github.com/google/learned_optimization/blob/aa15091066aa5b3f45e6b7f4bee1c41fb7d467a0/learned_optimization/tasks/fixed/rnn_lm.py#L142} for the impementation.

\subsection{Additional figures}
\begin{figure}[t]
    \centering
\begin{overpic}[width=1\textwidth]{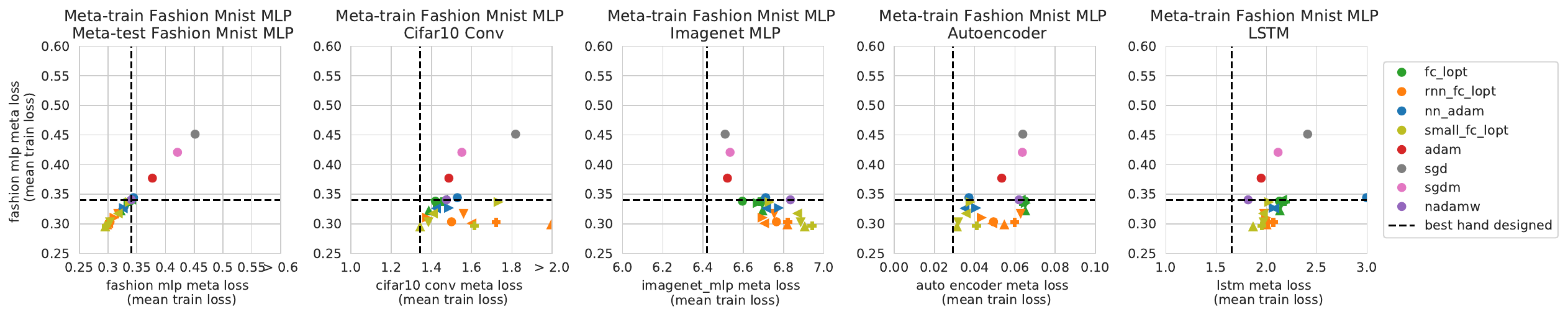}
\end{overpic}
\begin{overpic}[width=1\textwidth]{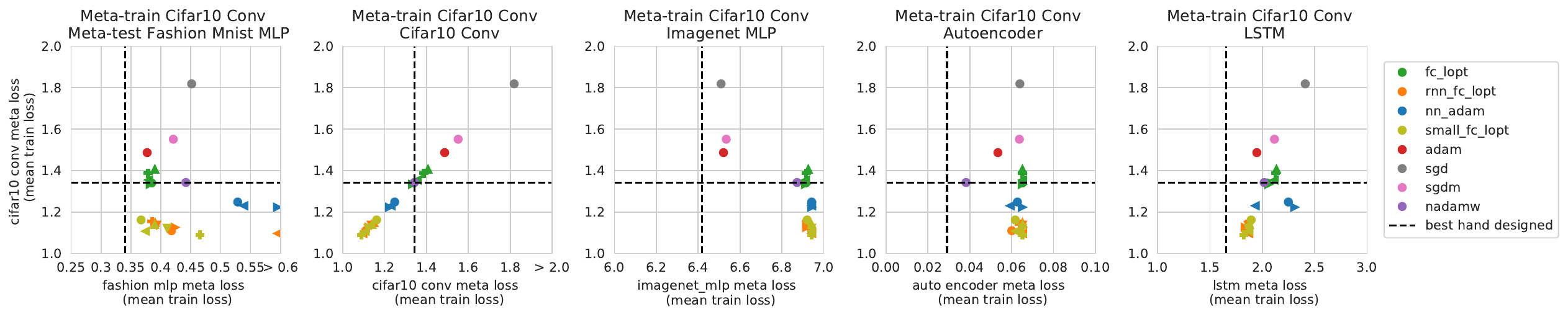}
\end{overpic}
    \caption{Further generalization results for more tasks. \textbf{Top row}: Meta-training on Fashion MNIST MLP and testing on 5 tasks. \textbf{Bottom row}: Meta-training on CIFAR-10 ConvNet and meta-testing on 5 tasks. \label{app_fig:meta-gen-tasks}}
\end{figure}

In this section we provide additional experimental results for meta-generalization similar to \sref{sec:outer_generalize}. First, in Figure \ref{app_fig:meta-gen-tasks} we show additional performance measurements on more held out tasks. As in \sref{sec:outer_generalize}, we see poor meta-generalization and high variability.

\begin{landscape}
\begin{figure}
    \centering
\begin{overpic}[width=1.3\textwidth]{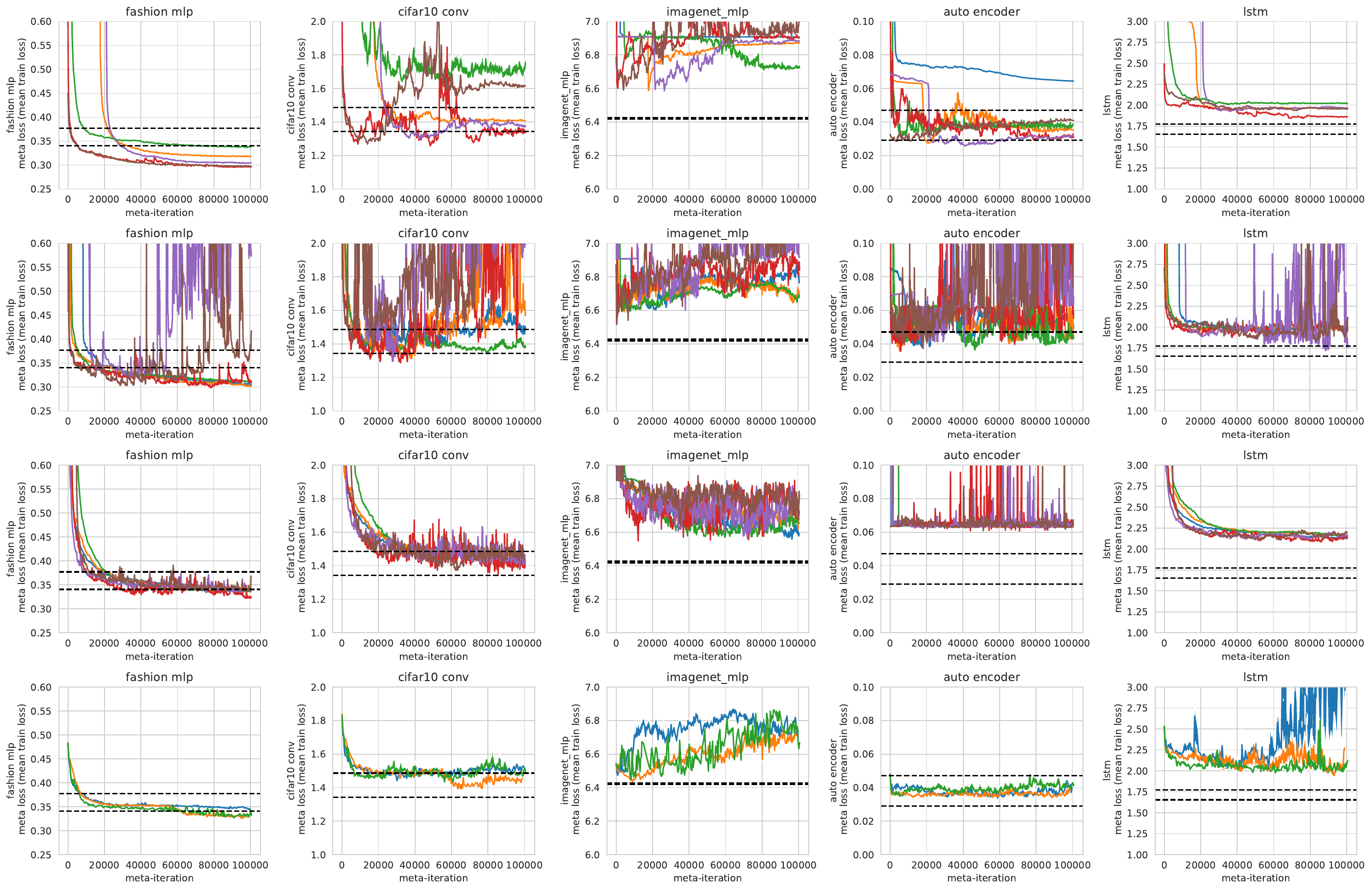}
\end{overpic}
    \caption{Meta-train and meta-test meta-training curves for different kinds of learned optimizers. In each row we show a different learned optimizer, with different colors denoting different random seed and or learning rate. From top to bottom we show small\_fc\_lopt, rnn\_lopt, fc\_lopt, nn\_adam. In black we show both Adam (top line) and NAdamW (bottom line) baselines tuned to the task being tested on.
    Error bars denote standard error across different random seeds when evaluating a given set of learned optimizer weights. Note how there is very little variation in evaluation of meta-loss, but large variation between different meta-iterations.
    In this pane we meta-train each learned optimizer on \textbf{Fashion MNIST MLP}. \label{curves:mlp}}
\end{figure}
\end{landscape}

\begin{landscape}
\begin{figure}
    \centering
\begin{overpic}[width=1.3\textwidth]{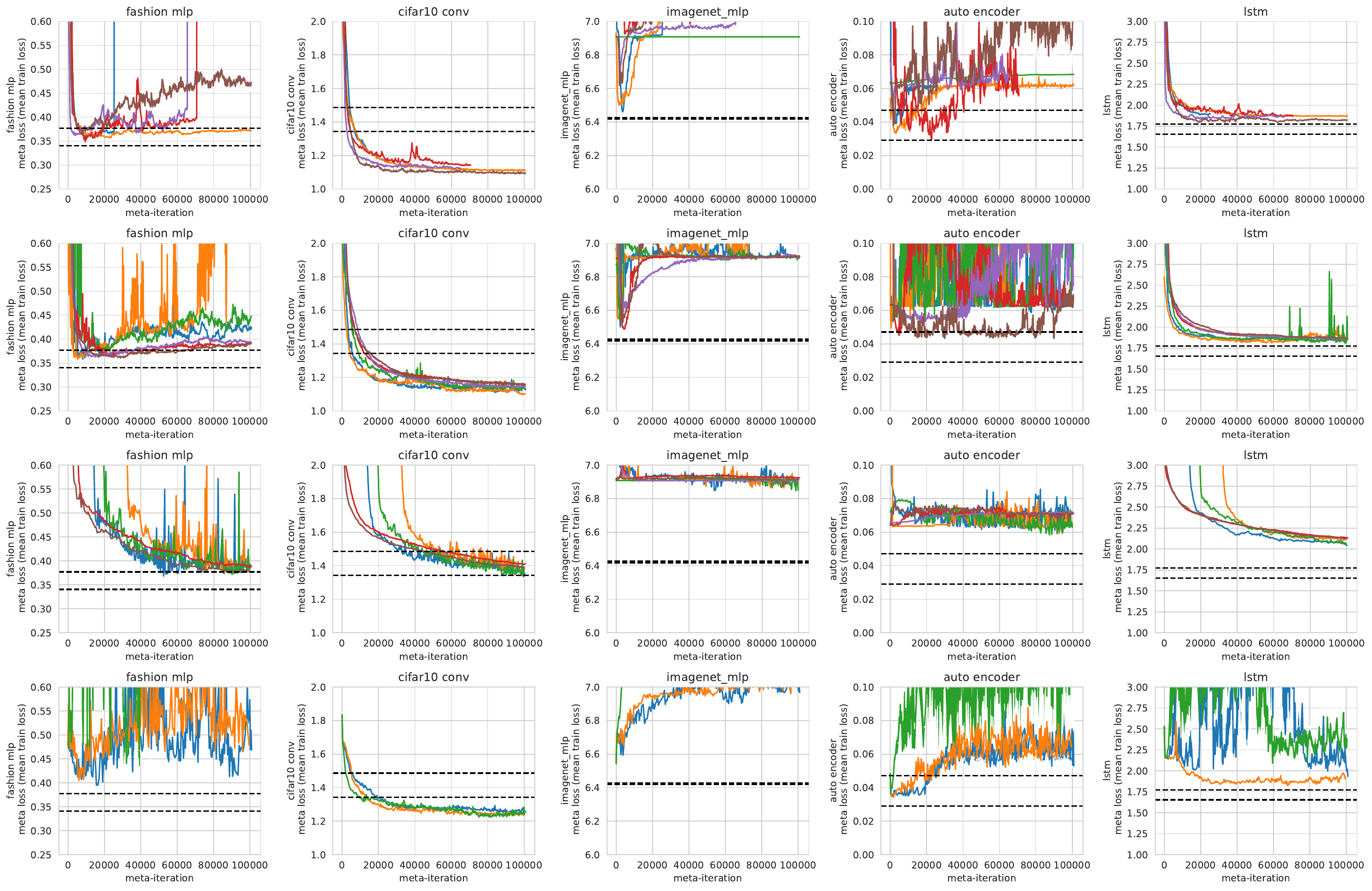}
% \put (3,21.0) {\textbf{\small(a)}}
% \put (33,21.0) {\textbf{\small(b)}}
% \put (62,21.0) {\textbf{\small(c)}}
\end{overpic}
    \caption{Meta-train and meta-test meta-training curves for different kinds of learned optimizers. In each row we show a different learned optimizer, with different colors denoting different random seed and or learning rate. From top to bottom we show small\_fc\_lopt, rnn\_lopt, fc\_lopt, nn\_adam.
    In black we show both Adam (top line) and NAdamW (bottom line) baselines tuned to the task being tested on.
    Error bars denote standard error across different random seeds when evaluating a given set of learned optimizer weights.
    Note how there is very little variation in evaluation of meta-loss, but large variation between different meta-iterations.
    In this pane we meta-train each learned optimizer on \textbf{CIFAR-10 ConvNet}. \label{curves:conv}}
\end{figure}
\end{landscape}

We plot evaluations over the course of meta-training for each different learned optimizer type and multiple random seeds in Figure~\ref{curves:mlp} when meta-training on the fashion Mnist MLP, and Figure~\ref{curves:conv} for the CIFAR-10 conv net. When meta-training and evaluating on the same distribution, we find stable evaluation loss. When evaluating on other kinds of tasks, we see wide variability in performance across both architecture, and even among different initializations of the learned optimizer weights holding all else fixed.
In some cases, such the learned optimizers switches between performing optimization on the target task, and diverging as shown by the rapid spikes in the meta-loss.

\begin{figure}
    \centering
\begin{overpic}[width=\textwidth]{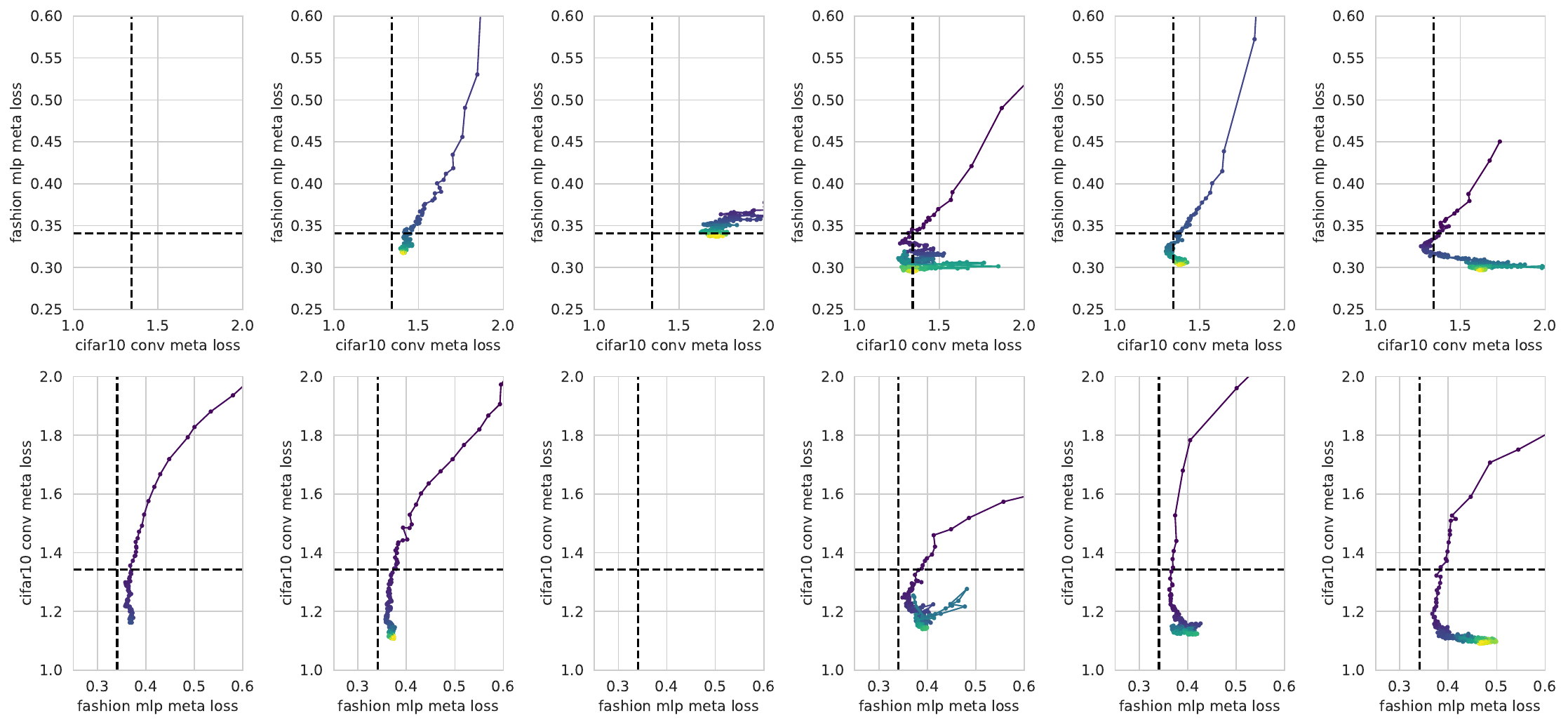}
% \put (3,21.0) {\textbf{\small(a)}}
% \put (33,21.0) {\textbf{\small(b)}}
% \put (62,21.0) {\textbf{\small(c)}}
\end{overpic}
    \caption{\textbf{small\_fc\_lopt Meta-generalization}: We show meta-training and meta-test performance plotted over the course of meta-training.
    In the top row we show meta-training on Fashion MNIST and meta-testing on the CIFAR-10 ConvNet.
    In the bottom we show meta-training on the CIFAR-10 ConvNet and meta-testing on Fashion MNIST MLP.
    Each column represents a random seed or learning rate -- the 3 left most columns are a smaller learning rate than the 3 right most columns.
    An empty plot indicates the model for the given seed did not converge.
    Purple is earlier in training, yellow is late in meta-training.
    We see meta-overfitting in all cases denoted by the C shaped curves. \label{curve:adafac}
    }
\end{figure}

\begin{figure}
    \centering
\begin{overpic}[width=\textwidth]{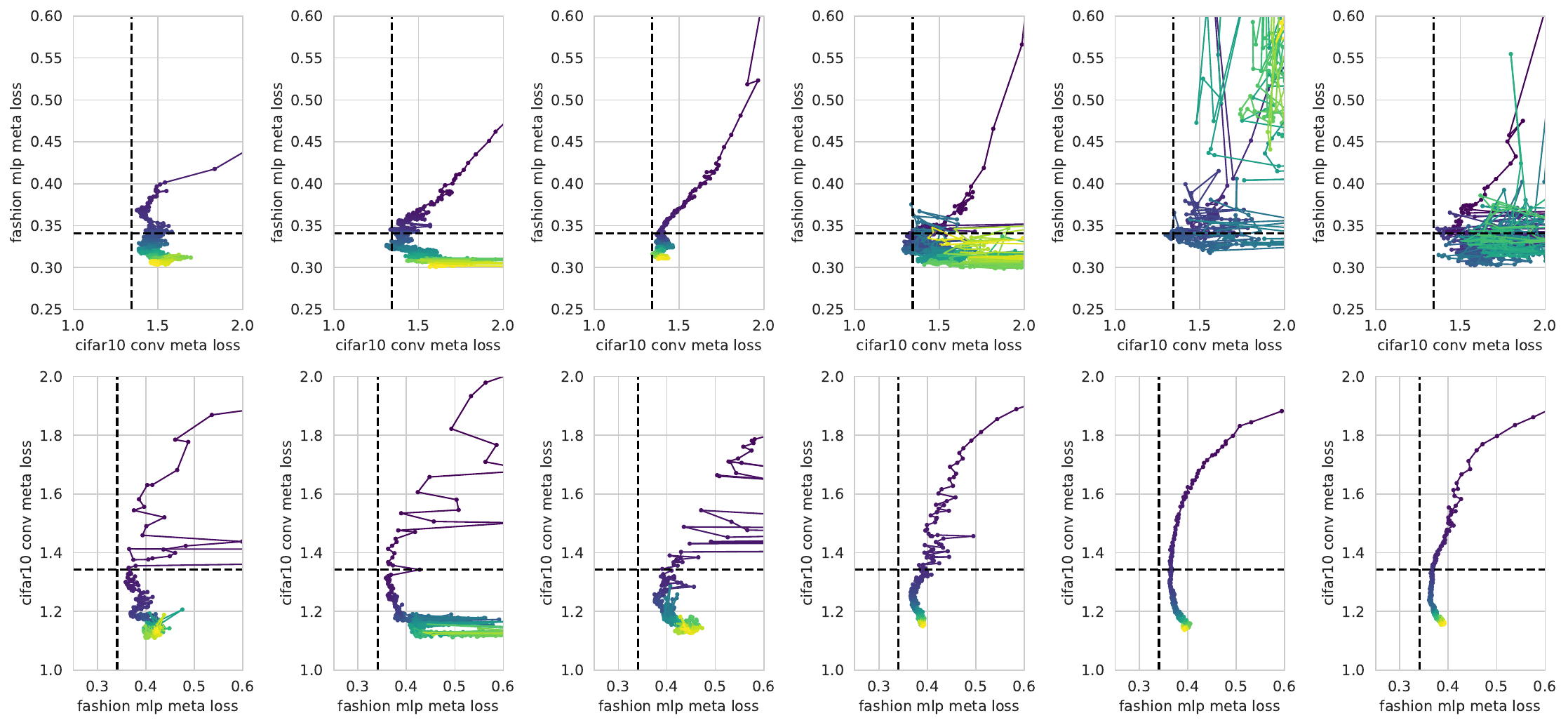}
% \put (3,21.0) {\textbf{\small(a)}}
% \put (33,21.0) {\textbf{\small(b)}}
% \put (62,21.0) {\textbf{\small(c)}}
\end{overpic}
    \caption{\textbf{rnn\_fc\_lopt Meta-generalization}: We show meta-training and meta-test performance plotted over the course of meta-training.
    In the top row we show meta-training on Fashion MNIST and meta-testing on the CIFAR-10 ConvNet.
    In the bottom we show meta-training on the CIFAR-10 ConvNet and meta-testing on Fashion MNIST.
    Each column represents a random seed or learning rate -- the 3 left most columns are a smaller learning rate than the 3 right most columns.
    An empty plot indicates the model for the given seed did not converge.
    Purple is earlier in training, yellow is late in meta-training.
    We see meta-overfitting in all cases denoted by the C shaped curves. \label{curve:rnn}
    }
\end{figure}

\begin{figure}
    \centering
\begin{overpic}[width=\textwidth]{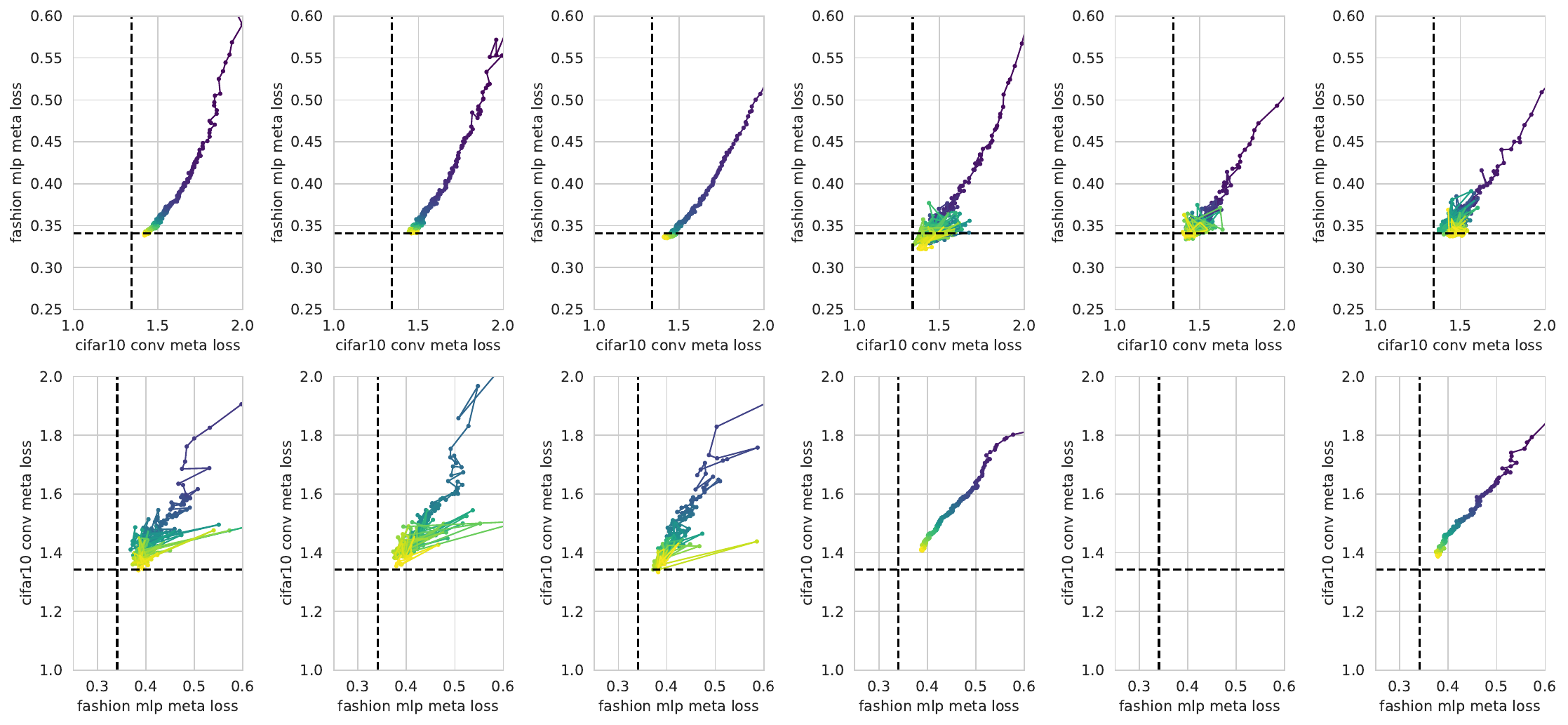}
% \put (3,21.0) {\textbf{\small(a)}}
% \put (33,21.0) {\textbf{\small(b)}}
% \put (62,21.0) {\textbf{\small(c)}}
\end{overpic}
    \caption{\textbf{fc\_lopt Meta-generalization}: We show meta-training and meta-test performance plotted over the course of meta-training. In the top row we show meta-training on Fashion MNIST and meta-testing on the CIFAR-10 ConvNet.
    In the bottom we show meta-training on the CIFAR-10 ConvNet and meta-testing on Fashion MNIST.
    Each column represents a random seed.
    Purple is earlier in training, yellow is late in meta-training.
    Here we see less meta-overfitting likely due to the learned optimizer not fully fitting the meta-training distribution. The empty plot denotes a learned optimizer which did not converge. \label{curve:mlplopt}
    }
\end{figure}

\begin{figure}
    \centering
\begin{overpic}[width=0.6\textwidth]{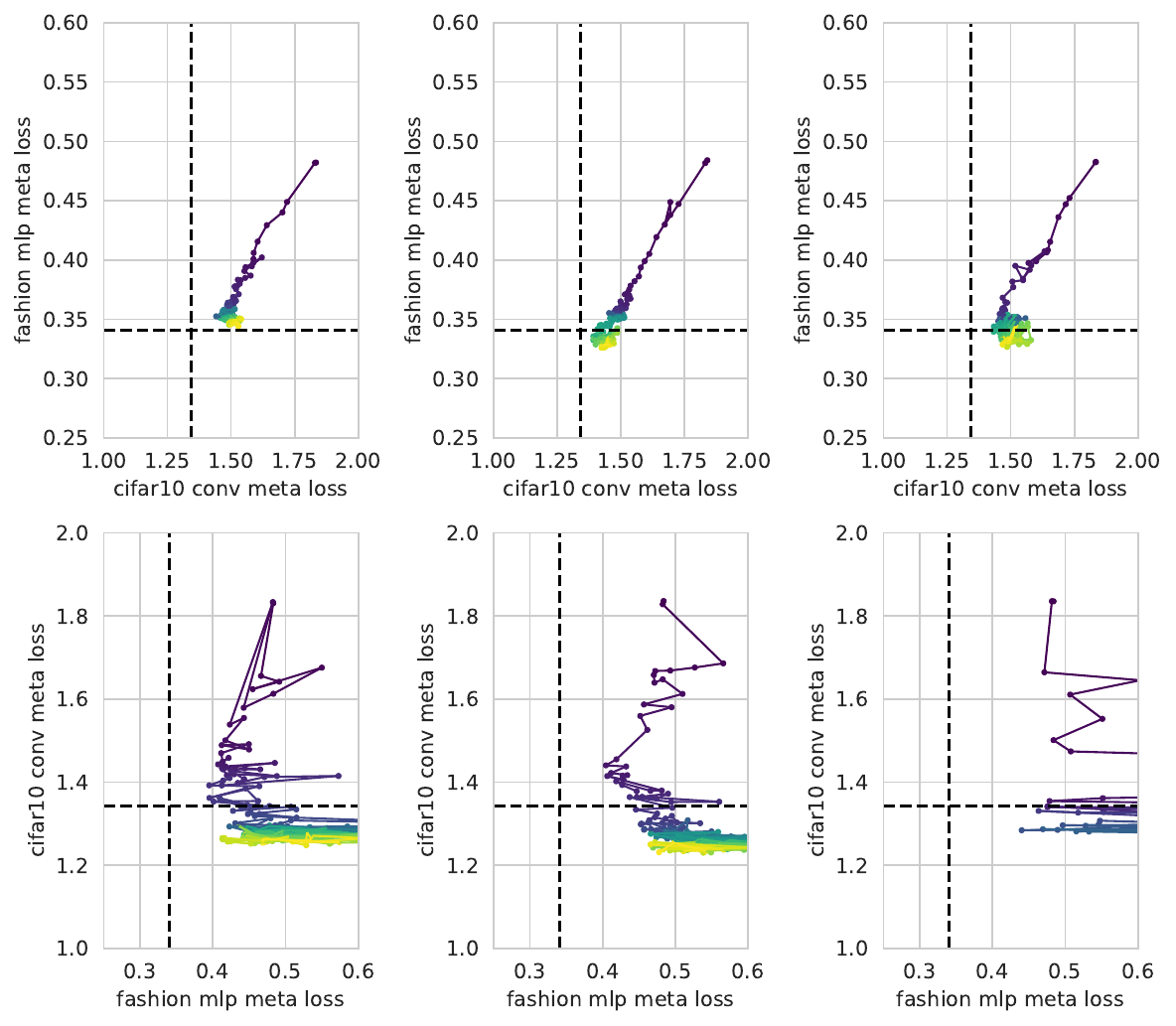}
% \put (3,21.0) {\textbf{\small(a)}}
% \put (33,21.0) {\textbf{\small(b)}}
% \put (62,21.0) {\textbf{\small(c)}}
\end{overpic}
    \caption{\textbf{nn\_adam Meta-generalization}: We show meta-training and meta-test performance plotted over the course of meta-training.
    In the top row we show meta-training on fashion MNIST and meta-testing on the CIFAR-10 ConvNet.
    In the bottom we show meta-training on the CIFAR-10 ConvNet and meta-testing on Fashion MNIST MLP.
    Each column represents a random seed. The fifth figure on the second row is empty as meta-training diverged.
    Purple is earlier in training, yellow is late in meta-training.\label{curve:nnadam}}
\end{figure}

Finally, we show an alternative plot of the same data discussed in the previous paragraph. This time, we plot meta-evaluation performance against meta-train performance. For each figure we show each learning rate, and each seed in a separate pane. We show the small\_fc\_lopt optimizer in Figure~\ref{curve:adafac}, the rnn\_fc\_lopt in Figure~\ref{curve:rnn}, the fc\_lopt in Figure~\ref{curve:mlplopt}, and nn\_adam in Figure~\ref{curve:nnadam}. Once again we find high variability across architecture, learning rate, and random seed. In these figures, meta-overfitting is highlighted by a "c" shaped curve -- meta-training performance continues to improve, but meta-evaluation performance gets worse after some point. Jagged lines / instability suggest a high sensitivity in performance on the evaluation task.

\clearpage

% \clearpage

\section{NAdamW update equations and search space} \label{app:nadamw_form}
For our NAdamW baseline, we use the same implementation, and search space described in \citet{metz2020using}. We repeat the functional form here for convenience.

\subsection{Update equations}
This optimizer architecture has 10 hyperparameters.
The base learning rate, $\alpha_{base}$, first and second moment momentum, $\beta_1$, $\beta_2$, the numerical stability term, $\epsilon$, $\ell_{2WD}$ $\ell_2$ regularization strength, $\ell_{2AdamW}$ AdamW style weight decay, and a boolean to switch between NAdam and Adam, $b_{\text{use nesterov}}$. The learning rate schedule is based off of a single cycle cosine decay with a warmup. It is controlled by 3 additional parameters -- $c_{\text{warmup}}$, $c_{\text{constant}}$, and $c_{\text{min learning rate mult}}$.

The learning rate is defined by:
\begin{align}
    u =& c_{\text{warmup}} T > t\\
    \alpha_{\text{decay\&constant}} =& (\alpha_{base} - c_{\text{min learning rate mult}})(0.5\\
        &\operatorname{cos}(t\pi / (T - c_{\text{constant}}))+ 0.5) + \\
        & c_{\text{min learning rate mult}}\\
    \alpha_{\text{warmup}} =& \frac{t}{(Tc_{\text{warmup)}}}\\
    \alpha =& (1 - u)\alpha_{\text{decay\&constant}} + u \alpha_{\text{warm}}
\end{align}

The update equations of NAdamW follow.
\begin{align}
    \phi^{(0)} =& \text{problem specified random initialization}\\
    m^{(0)} =& 0 \\
    v^{(0)} =& 0 \\
    g^{(t)} =& \frac{d}{d\phi^{(t)}}(f(x; \phi^{(t)}) + \ell_{2wd}||\phi^{(t)}||^2_2 )\\
    m^{(t)} =& \beta_1 m^{(t-1)} + g^{(t)}(1 - \beta_1) \\
    v^{(t)} =& \beta_2 v^{(t-1)} + (g^{(t)})^2(1 - \beta_2) \\
    \hat{m}^{(t)} =& \dfrac{m^{(t)}}{1-\beta_1^{t+1}} \\
    \hat{v}^{(t)} =& \frac{v^{(t)}}{1-\beta_2^{t+1}} \\
    u^{(t)}_{\text{heavy ball}} =& \dfrac{\hat{m}^{(t)}}{\sqrt{\hat{v}^{(t)}} + \epsilon}\\
    u^{(t)}_{\text{nesterov}} =& \dfrac{\beta_1 \hat{m}^{(t)} + (1 - \beta_1)g^{(t)}} {\sqrt{\hat{v}^{(t)}} + \epsilon} \\
    \phi^{(t+1)} =& \phi^{(t)} - (1-b_{\text{use nesterov}}) \alpha u^{(t)}_{{\text{heavy ball}}}  +\\
     & b_{\text{use nesterov}} \alpha u^{(t)}_{\text{nesterov}} - \alpha \ell_{2AdamW} \phi^{(t)}
\end{align}

\subsection{hyperparameter search space} \label{app:nadamw_search}
The initial learning rate, $\alpha_{base}$ is sampled from log space between $1e-5$ and $1.0$. $1-\beta_1$ is sampled logarithmically between $1e-3$, and $1.0$. $1-\beta_2$ is sampled between $1e-5$, and $1.0$. $\epsilon$ is sampled logarithmically between $1e-8$ and $1e4$. We sample using nesterov ($b_{\text{use nesterov}}$) 50\% of the time. 
We sample $\ell_{2WD}$ and $\ell_{2AdamW}$ logarithmically between $1e-5$ and $1e-1$. Equal probabilities of a third we either use both terms, zero out $\ell_{2WD}$, or zero out $\ell_{2AdamW}$.
With 50\% probability we use a nonzero min learning rate multiplier sampled logarithmically between $1e-5$ and $1.0$.
With 50\% probability we sample the warm up fraction, $c_{\text{warmup}}$ between 1e-5 and 1e-1, otherwise it is set to zero.
Finally, we uniformly sample the amount of time the learning rate is held constant ($c_{\text{constant}}$) between 0 and 1.
\end{document}

%% file: math_commands.tex
\usepackage{amsmath,amsfonts,bm}

\def\eqref#1{equation~\ref{#1}}
\def\1{\bm{1}}

\DeclareMathAlphabet{\mathsfit}{\encodingdefault}{\sfdefault}{m}{sl}
\SetMathAlphabet{\mathsfit}{bold}{\encodingdefault}{\sfdefault}{bx}{n}

%% file: main.bbl
\begin{thebibliography}{81}
\providecommand{\natexlab}[1]{#1}
\providecommand{\url}[1]{\texttt{#1}}
\expandafter\ifx\csname urlstyle\endcsname\relax
  \providecommand{\doi}[1]{doi: #1}\else
  \providecommand{\doi}{doi: \begingroup \urlstyle{rm}\Url}\fi

\bibitem[Akiba et~al.(2019)Akiba, Sano, Yanase, Ohta, and Koyama]{optuna_2019}
Takuya Akiba, Shotaro Sano, Toshihiko Yanase, Takeru Ohta, and Masanori Koyama.
\newblock Optuna: A next-generation hyperparameter optimization framework.
\newblock In \emph{Proceedings of the 25rd {ACM} {SIGKDD} International
  Conference on Knowledge Discovery and Data Mining}, 2019.

\bibitem[Almeida et~al.(2021)Almeida, Winter, Tang, and
  Zaremba]{almeida2021generalizable}
Diogo Almeida, Clemens Winter, Jie Tang, and Wojciech Zaremba.
\newblock A generalizable approach to learning optimizers.
\newblock \emph{arXiv preprint arXiv:2106.00958}, 2021.

\bibitem[Andrychowicz et~al.(2016)Andrychowicz, Denil, Gomez, Hoffman, Pfau,
  Schaul, and de~Freitas]{andrychowicz2016learning}
Marcin Andrychowicz, Misha Denil, Sergio Gomez, Matthew~W Hoffman, David Pfau,
  Tom Schaul, and Nando de~Freitas.
\newblock Learning to learn by gradient descent by gradient descent.
\newblock In \emph{Advances in Neural Information Processing Systems}, pages
  3981--3989, 2016.

\bibitem[Anil et~al.(2019)Anil, Gupta, Koren, and Singer]{anil2019memory}
Rohan Anil, Vineet Gupta, Tomer Koren, and Yoram Singer.
\newblock Memory-efficient adaptive optimization.
\newblock \emph{arXiv preprint arXiv:1901.11150}, 2019.

\bibitem[Anil et~al.(2020)Anil, Gupta, Koren, Regan, and
  Singer]{anil2020second}
Rohan Anil, Vineet Gupta, Tomer Koren, Kevin Regan, and Yoram Singer.
\newblock Second order optimization made practical.
\newblock \emph{arXiv preprint arXiv:2002.09018}, 2020.

\bibitem[Bello et~al.(2017)Bello, Zoph, Vasudevan, and Le]{Bello17}
Irwan Bello, Barret Zoph, Vijay Vasudevan, and Quoc Le.
\newblock Neural optimizer search with reinforcement learning.
\newblock 2017.
\newblock URL \url{https://arxiv.org/pdf/1709.07417.pdf}.

\bibitem[Bengio et~al.(1992)Bengio, Bengio, Cloutier, and
  Gecsei]{bengio1992optimization}
Samy Bengio, Yoshua Bengio, Jocelyn Cloutier, and Jan Gecsei.
\newblock On the optimization of a synaptic learning rule.
\newblock In \emph{Preprints Conf. Optimality in Artificial and Biological
  Neural Networks}, pages 6--8. Univ. of Texas, 1992.

\bibitem[Bergstra and Bengio(2012)]{bergstra2012random}
James Bergstra and Yoshua Bengio.
\newblock Random search for hyper-parameter optimization.
\newblock \emph{Journal of Machine Learning Research}, 13\penalty0
  (Feb):\penalty0 281--305, 2012.

\bibitem[Bergstra et~al.(2011)Bergstra, Bardenet, Bengio, and
  K{\'e}gl]{bergstra2011algorithms}
James Bergstra, R{\'e}mi Bardenet, Yoshua Bengio, and Bal{\'a}zs K{\'e}gl.
\newblock Algorithms for hyper-parameter optimization.
\newblock In \emph{25th annual conference on neural information processing
  systems (NIPS 2011)}, volume~24. Neural Information Processing Systems
  Foundation, 2011.

\bibitem[Bottou(2010)]{bottou2010large}
L{\'e}on Bottou.
\newblock Large-scale machine learning with stochastic gradient descent.
\newblock In \emph{Proceedings of COMPSTAT'2010}, pages 177--186. Springer,
  2010.

\bibitem[Bradbury et~al.(2018)Bradbury, Frostig, Hawkins, Johnson, Leary,
  Maclaurin, Necula, Paszke, Vander{P}las, Wanderman-{M}ilne, and
  Zhang]{jax2018github}
James Bradbury, Roy Frostig, Peter Hawkins, Matthew~James Johnson, Chris Leary,
  Dougal Maclaurin, George Necula, Adam Paszke, Jake Vander{P}las, Skye
  Wanderman-{M}ilne, and Qiao Zhang.
\newblock {JAX}: composable transformations of {P}ython+{N}um{P}y programs,
  2018.
\newblock URL \url{http://github.com/google/jax}.

\bibitem[Broyden(1970)]{broyden1970convergence}
Charles~George Broyden.
\newblock The convergence of a class of double-rank minimization algorithms 1.
  general considerations.
\newblock \emph{IMA Journal of Applied Mathematics}, 6\penalty0 (1):\penalty0
  76--90, 1970.

\bibitem[Cao et~al.(2019)Cao, Chen, Wang, and Shen]{cao2019learning}
Yue Cao, Tianlong Chen, Zhangyang Wang, and Yang Shen.
\newblock Learning to optimize in swarms.
\newblock \emph{arXiv preprint arXiv:1911.03787}, 2019.

\bibitem[Chelba et~al.(2013)Chelba, Mikolov, Schuster, Ge, Brants, and
  Koehn]{DBLP:journals/corr/ChelbaMSGBK13}
Ciprian Chelba, Tomas Mikolov, Mike Schuster, Qi~Ge, Thorsten Brants, and
  Phillipp Koehn.
\newblock One billion word benchmark for measuring progress in statistical
  language modeling.
\newblock \emph{CoRR}, abs/1312.3005, 2013.
\newblock URL \url{http://arxiv.org/abs/1312.3005}.

\bibitem[Chen et~al.(2020)Chen, Zhang, Jingyang, Chang, Liu, Amini, and
  Wang]{chen2020training}
Tianlong Chen, Weiyi Zhang, Zhou Jingyang, Shiyu Chang, Sijia Liu, Lisa Amini,
  and Zhangyang Wang.
\newblock Training stronger baselines for learning to optimize.
\newblock \emph{Advances in Neural Information Processing Systems}, 33, 2020.

\bibitem[Chen et~al.(2016)Chen, Hoffman, Colmenarejo, Denil, Lillicrap,
  Botvinick, and de~Freitas]{chen2016learning}
Yutian Chen, Matthew~W Hoffman, Sergio~G{\'o}mez Colmenarejo, Misha Denil,
  Timothy~P Lillicrap, Matt Botvinick, and Nando de~Freitas.
\newblock Learning to learn without gradient descent by gradient descent.
\newblock \emph{arXiv preprint arXiv:1611.03824}, 2016.

\bibitem[Daniel et~al.(2016)Daniel, Taylor, and Nowozin]{daniel2016learning}
Christian Daniel, Jonathan Taylor, and Sebastian Nowozin.
\newblock Learning step size controllers for robust neural network training.
\newblock In \emph{Thirtieth AAAI Conference on Artificial Intelligence}, 2016.

\bibitem[Dennis and Mor{\'e}(1977)]{dennis1977quasi}
John~E Dennis, Jr and Jorge~J Mor{\'e}.
\newblock Quasi-newton methods, motivation and theory.
\newblock \emph{SIAM review}, 19\penalty0 (1):\penalty0 46--89, 1977.

\bibitem[Dozat(2016)]{dozat2016incorporating}
Timothy Dozat.
\newblock Incorporating nesterov momentum into adam.
\newblock 2016.

\bibitem[Duchi et~al.(2011)Duchi, Hazan, and Singer]{duchi2011adaptive}
John Duchi, Elad Hazan, and Yoram Singer.
\newblock Adaptive subgradient methods for online learning and stochastic
  optimization.
\newblock \emph{Journal of machine learning research}, 12\penalty0 (7), 2011.

\bibitem[Finn et~al.(2017)Finn, Abbeel, and Levine]{finn2017model}
Chelsea Finn, Pieter Abbeel, and Sergey Levine.
\newblock Model-agnostic meta-learning for fast adaptation of deep networks.
\newblock \emph{arXiv preprint arXiv:1703.03400}, 2017.

\bibitem[Golovin et~al.(2017)Golovin, Solnik, Moitra, Kochanski, Karro, and
  Sculley]{golovin2017google}
Daniel Golovin, Benjamin Solnik, Subhodeep Moitra, Greg Kochanski, John Karro,
  and D~Sculley.
\newblock Google vizier: A service for black-box optimization.
\newblock In \emph{Proceedings of the 23rd ACM SIGKDD international conference
  on knowledge discovery and data mining}, pages 1487--1495, 2017.

\bibitem[Gong et~al.(2018)Gong, Li, and Hern{\'a}ndez-Lobato]{gong2018meta}
Wenbo Gong, Yingzhen Li, and Jos{\'e}~Miguel Hern{\'a}ndez-Lobato.
\newblock Meta-learning for stochastic gradient mcmc.
\newblock \emph{arXiv preprint arXiv:1806.04522}, 2018.

\bibitem[Gupta et~al.(2018)Gupta, Koren, and Singer]{gupta2018shampoo}
Vineet Gupta, Tomer Koren, and Yoram Singer.
\newblock Shampoo: Preconditioned stochastic tensor optimization.
\newblock In \emph{International Conference on Machine Learning}, pages
  1842--1850. PMLR, 2018.

\bibitem[Hansen(2016)]{hansen2016using}
Samantha Hansen.
\newblock Using deep q-learning to control optimization hyperparameters.
\newblock \emph{arXiv preprint arXiv:1602.04062}, 2016.

\bibitem[He et~al.(2016)He, Zhang, Ren, and Sun]{he2016deep}
Kaiming He, Xiangyu Zhang, Shaoqing Ren, and Jian Sun.
\newblock Deep residual learning for image recognition.
\newblock In \emph{Proceedings of the IEEE conference on computer vision and
  pattern recognition}, pages 770--778, 2016.

\bibitem[Hennigan et~al.(2020)Hennigan, Cai, Norman, and
  Babuschkin]{haiku2020github}
Tom Hennigan, Trevor Cai, Tamara Norman, and Igor Babuschkin.
\newblock {H}aiku: {S}onnet for {JAX}, 2020.
\newblock URL \url{http://github.com/deepmind/dm-haiku}.

\bibitem[Hochreiter and Schmidhuber(1997)]{hochreiter1997long}
Sepp Hochreiter and J{\"u}rgen Schmidhuber.
\newblock Long short-term memory.
\newblock \emph{Neural computation}, 9\penalty0 (8):\penalty0 1735--1780, 1997.

\bibitem[Hochreiter et~al.(2001)Hochreiter, Younger, and
  Conwell]{hochreiter2001learning}
Sepp Hochreiter, A~Steven Younger, and Peter~R Conwell.
\newblock Learning to learn using gradient descent.
\newblock In \emph{International Conference on Artificial Neural Networks},
  pages 87--94. Springer, 2001.

\bibitem[Hunter(2007)]{matplotlib}
J.~D. Hunter.
\newblock Matplotlib: A 2d graphics environment.
\newblock \emph{Computing in Science \& Engineering}, 9\penalty0 (3):\penalty0
  90--95, 2007.
\newblock \doi{10.1109/MCSE.2007.55}.

\bibitem[Kingma and Ba(2014)]{kingma2014adam}
Diederik~P Kingma and Jimmy Ba.
\newblock Adam: A method for stochastic optimization.
\newblock \emph{arXiv preprint arXiv:1412.6980}, 2014.

\bibitem[Krizhevsky et~al.(2009)Krizhevsky, Nair, and
  Hinton]{krizhevsky2009cifar}
Alex Krizhevsky, Vinod Nair, and Geoffrey Hinton.
\newblock Cifar-10 and cifar-100 datasets.
\newblock \emph{URl: https://www. cs. toronto. edu/kriz/cifar. html}, 6, 2009.

\bibitem[Le et~al.(2011)Le, Ngiam, Coates, Lahiri, Prochnow, and
  Ng]{le2011optimization}
Quoc~V Le, Jiquan Ngiam, Adam Coates, Ahbik Lahiri, Bobby Prochnow, and
  Andrew~Y Ng.
\newblock On optimization methods for deep learning.
\newblock In \emph{ICML}, 2011.

\bibitem[LeCun et~al.(2015)LeCun, Bengio, and Hinton]{lecun2015deep}
Yann LeCun, Yoshua Bengio, and Geoffrey Hinton.
\newblock Deep learning.
\newblock \emph{nature}, 521\penalty0 (7553):\penalty0 436--444, 2015.

\bibitem[Levy et~al.(2017)Levy, Hoffman, and
  Sohl-Dickstein]{levy2017generalizing}
Daniel Levy, Matthew~D Hoffman, and Jascha Sohl-Dickstein.
\newblock Generalizing hamiltonian monte carlo with neural networks.
\newblock \emph{arXiv preprint arXiv:1711.09268}, 2017.

\bibitem[Li and Malik(2017{\natexlab{a}})]{li2016learning}
Ke~Li and Jitendra Malik.
\newblock Learning to optimize.
\newblock \emph{International Conference on Learning Representations},
  2017{\natexlab{a}}.

\bibitem[Li and Malik(2017{\natexlab{b}})]{li2017learning}
Ke~Li and Jitendra Malik.
\newblock Learning to optimize neural nets.
\newblock \emph{arXiv preprint arXiv:1703.00441}, 2017{\natexlab{b}}.

\bibitem[Liu and Nocedal(1989)]{liu1989limited}
Dong~C Liu and Jorge Nocedal.
\newblock On the limited memory bfgs method for large scale optimization.
\newblock \emph{Mathematical programming}, 45\penalty0 (1):\penalty0 503--528,
  1989.

\bibitem[Liu et~al.(2019)Liu, Jiang, He, Chen, Liu, Gao, and
  Han]{liu2019variance}
Liyuan Liu, Haoming Jiang, Pengcheng He, Weizhu Chen, Xiaodong Liu, Jianfeng
  Gao, and Jiawei Han.
\newblock On the variance of the adaptive learning rate and beyond.
\newblock \emph{arXiv preprint arXiv:1908.03265}, 2019.

\bibitem[Loshchilov and Hutter(2017)]{loshchilov2017decoupled}
Ilya Loshchilov and Frank Hutter.
\newblock Decoupled weight decay regularization.
\newblock \emph{arXiv preprint arXiv:1711.05101}, 2017.

\bibitem[Lucas et~al.(2018)Lucas, Zemel, and Grosse]{lucas2018aggregated}
James Lucas, Richard Zemel, and Roger Grosse.
\newblock Aggregated momentum: Stability through passive damping.
\newblock \emph{arXiv preprint arXiv:1804.00325}, 2018.

\bibitem[Lv et~al.(2017)Lv, Jiang, and Li]{lv2017learning}
Kaifeng Lv, Shunhua Jiang, and Jian Li.
\newblock Learning gradient descent: Better generalization and longer horizons.
\newblock \emph{arXiv preprint arXiv:1703.03633}, 2017.

\bibitem[Maclaurin et~al.(2015)Maclaurin, Duvenaud, and
  Adams]{maclaurin2015gradient}
Dougal Maclaurin, David Duvenaud, and Ryan Adams.
\newblock Gradient-based hyperparameter optimization through reversible
  learning.
\newblock In \emph{International Conference on Machine Learning}, pages
  2113--2122, 2015.

\bibitem[Maheswaranathan et~al.(2019)Maheswaranathan, Metz, Tucker, Choi, and
  Sohl-Dickstein]{maheswaranathan2019guided}
Niru Maheswaranathan, Luke Metz, George Tucker, Dami Choi, and Jascha
  Sohl-Dickstein.
\newblock Guided evolutionary strategies: Augmenting random search with
  surrogate gradients.
\newblock In \emph{International Conference on Machine Learning}, pages
  4264--4273. PMLR, 2019.

\bibitem[Maheswaranathan et~al.(2020)Maheswaranathan, Sussillo, Metz, Sun, and
  Sohl-Dickstein]{maheswaranathan2020reverse}
Niru Maheswaranathan, David Sussillo, Luke Metz, Ruoxi Sun, and Jascha
  Sohl-Dickstein.
\newblock Reverse engineering learned optimizers reveals known and novel
  mechanisms.
\newblock \emph{arXiv preprint arXiv:2011.02159}, 2020.

\bibitem[Martens and Grosse(2015)]{martens2015optimizing}
James Martens and Roger Grosse.
\newblock Optimizing neural networks with kronecker-factored approximate
  curvature.
\newblock In \emph{International conference on machine learning}, pages
  2408--2417. PMLR, 2015.

\bibitem[Merchant et~al.(2021)Merchant, Metz, Schoenholz, and Cubuk]{learn2hop}
Amil Merchant, Luke Metz, Sam Schoenholz, and Ekin~Dogus Cubuk.
\newblock Learn2hop: Learned optimization on rough landscapes.
\newblock \emph{ICML}, 2021.

\bibitem[Metz et~al.(2018)Metz, Maheswaranathan, Cheung, and
  Sohl-Dickstein]{metz2018learning}
Luke Metz, Niru Maheswaranathan, Brian Cheung, and Jascha Sohl-Dickstein.
\newblock Learning unsupervised learning rules.
\newblock \emph{arXiv preprint arXiv:1804.00222}, 2018.

\bibitem[Metz et~al.(2019{\natexlab{a}})Metz, Maheswaranathan, Nixon, Freeman,
  and Sohl-Dickstein]{metz2019understanding}
Luke Metz, Niru Maheswaranathan, Jeremy Nixon, Daniel Freeman, and Jascha
  Sohl-Dickstein.
\newblock Understanding and correcting pathologies in the training of learned
  optimizers.
\newblock In \emph{International Conference on Machine Learning}, pages
  4556--4565, 2019{\natexlab{a}}.

\bibitem[Metz et~al.(2019{\natexlab{b}})Metz, Maheswaranathan, Shlens,
  Sohl-Dickstein, and Cubuk]{metz2019using}
Luke Metz, Niru Maheswaranathan, Jonathon Shlens, Jascha Sohl-Dickstein, and
  Ekin~D Cubuk.
\newblock Using learned optimizers to make models robust to input noise.
\newblock \emph{arXiv preprint arXiv:1906.03367}, 2019{\natexlab{b}}.

\bibitem[Metz et~al.(2020{\natexlab{a}})Metz, Maheswaranathan, Freeman, Poole,
  and Sohl-Dickstein]{metz2020tasks}
Luke Metz, Niru Maheswaranathan, C~Daniel Freeman, Ben Poole, and Jascha
  Sohl-Dickstein.
\newblock Tasks, stability, architecture, and compute: Training more effective
  learned optimizers, and using them to train themselves.
\newblock \emph{arXiv preprint arXiv:2009.11243}, 2020{\natexlab{a}}.

\bibitem[Metz et~al.(2020{\natexlab{b}})Metz, Maheswaranathan, Sun, Freeman,
  Poole, and Sohl-Dickstein]{metz2020using}
Luke Metz, Niru Maheswaranathan, Ruoxi Sun, C~Daniel Freeman, Ben Poole, and
  Jascha Sohl-Dickstein.
\newblock Using a thousand optimization tasks to learn hyperparameter search
  strategies.
\newblock \emph{arXiv preprint arXiv:2002.11887}, 2020{\natexlab{b}}.

\bibitem[Metz et~al.(2021)Metz, Freeman, Maheswaranathan, and
  Sohl-Dickstein]{metz2021training}
Luke Metz, C~Daniel Freeman, Niru Maheswaranathan, and Jascha Sohl-Dickstein.
\newblock Training learned optimizers with randomly initialized learned
  optimizers.
\newblock \emph{arXiv preprint arXiv:2101.07367}, 2021.

\bibitem[Nesterov(1983)]{nesterov1983method}
Yurii Nesterov.
\newblock A method for unconstrained convex minimization problem with the rate
  of convergence o (1/k\^{} 2).
\newblock In \emph{Doklady AN USSR}, volume 269, pages 543--547, 1983.

\bibitem[Newman(1998)]{newman1998new}
Peter Newman.
\newblock \emph{The new Palgrave dictionary of economics and the law}.
\newblock Springer, 1998.

\bibitem[Norrie et~al.(2020)Norrie, Patil, Yoon, Kurian, Li, Laudon, Young,
  Jouppi, and Patterson]{norrie2020google}
Thomas Norrie, Nishant Patil, Doe~Hyun Yoon, George Kurian, Sheng Li, James
  Laudon, Cliff Young, Norman~P Jouppi, and David Patterson.
\newblock Google’s training chips revealed: Tpuv2 and tpuv3.
\newblock In \emph{2020 IEEE Hot Chips 32 Symposium (HCS)}, pages 1--70. IEEE
  Computer Society, 2020.

\bibitem[Polyak(1964)]{polyak1964some}
Boris~T Polyak.
\newblock Some methods of speeding up the convergence of iteration methods.
\newblock \emph{Ussr computational mathematics and mathematical physics},
  4\penalty0 (5):\penalty0 1--17, 1964.

\bibitem[Ravi and Larochelle(2016)]{ravi2016optimization}
Sachin Ravi and Hugo Larochelle.
\newblock Optimization as a model for few-shot learning.
\newblock \emph{International Conference on Learning Representations}, 2016.

\bibitem[Reddi et~al.(2018)Reddi, Zaheer, Sachan, Kale, and
  Kumar]{reddi2018adaptive}
S~Reddi, Manzil Zaheer, Devendra Sachan, Satyen Kale, and Sanjiv Kumar.
\newblock Adaptive methods for nonconvex optimization.
\newblock In \emph{Proceeding of 32nd Conference on Neural Information
  Processing Systems (NIPS 2018)}, 2018.

\bibitem[Robbins and Monro(1951)]{robbins1951stochastic}
Herbert Robbins and Sutton Monro.
\newblock A stochastic approximation method.
\newblock \emph{The annals of mathematical statistics}, pages 400--407, 1951.

\bibitem[Ruder(2016)]{ruder2016overview}
Sebastian Ruder.
\newblock An overview of gradient descent optimization algorithms.
\newblock \emph{arXiv preprint arXiv:1609.04747}, 2016.

\bibitem[Runarsson and Jonsson(2000)]{runarsson2000evolution}
Thomas~Philip Runarsson and Magnus~Thor Jonsson.
\newblock Evolution and design of distributed learning rules.
\newblock In \emph{Combinations of Evolutionary Computation and Neural
  Networks, 2000 IEEE Symposium on}, pages 59--63. IEEE, 2000.

\bibitem[Schmidt et~al.(2020)Schmidt, Schneider, and
  Hennig]{schmidt2020descending}
Robin~M Schmidt, Frank Schneider, and Philipp Hennig.
\newblock Descending through a crowded valley--benchmarking deep learning
  optimizers.
\newblock \emph{arXiv preprint arXiv:2007.01547}, 2020.

\bibitem[Shazeer and Stern(2018)]{shazeer2018adafactor}
Noam Shazeer and Mitchell Stern.
\newblock Adafactor: Adaptive learning rates with sublinear memory cost.
\newblock \emph{arXiv preprint arXiv:1804.04235}, 2018.

\bibitem[Shen et~al.(2021)Shen, Chen, Heaton, Chen, Liu, Yin, and
  Wang]{shen2021learning}
Jiayi Shen, Xiaohan Chen, Howard Heaton, Tianlong Chen, Jialin Liu, Wotao Yin,
  and Zhangyang Wang.
\newblock Learning a minimax optimizer: A pilot study.
\newblock In \emph{International Conference on Learning Representations
  (ICLR)}, 2021.

\bibitem[Simonyan and Zisserman(2014)]{simonyan2014very}
Karen Simonyan and Andrew Zisserman.
\newblock Very deep convolutional networks for large-scale image recognition.
\newblock \emph{arXiv preprint arXiv:1409.1556}, 2014.

\bibitem[Snoek et~al.(2012)Snoek, Larochelle, and Adams]{snoek2012practical}
Jasper Snoek, Hugo Larochelle, and Ryan~P Adams.
\newblock Practical bayesian optimization of machine learning algorithms.
\newblock In \emph{Advances in neural information processing systems}, pages
  2951--2959, 2012.

\bibitem[Tan and Le(2019)]{tan2019efficientnet}
Mingxing Tan and Quoc Le.
\newblock Efficientnet: Rethinking model scaling for convolutional neural
  networks.
\newblock In \emph{International Conference on Machine Learning}, pages
  6105--6114. PMLR, 2019.

\bibitem[Tieleman and Hinton(2012)]{tieleman2012lecture}
Tijmen Tieleman and Geoffrey Hinton.
\newblock Lecture 6.5-rmsprop: Divide the gradient by a running average of its
  recent magnitude.
\newblock \emph{COURSERA: Neural networks for machine learning}, 4\penalty0
  (2):\penalty0 26--31, 2012.

\bibitem[Van Der~Walt et~al.(2011)Van Der~Walt, Colbert, and
  Varoquaux]{van2011numpy}
Stefan Van Der~Walt, S~Chris Colbert, and Gael Varoquaux.
\newblock The numpy array: a structure for efficient numerical computation.
\newblock \emph{Computing in Science \& Engineering}, 13\penalty0 (2):\penalty0
  22, 2011.

\bibitem[Vaswani et~al.(2017)Vaswani, Shazeer, Parmar, Uszkoreit, Jones, Gomez,
  Kaiser, and Polosukhin]{vaswani2017attention}
Ashish Vaswani, Noam Shazeer, Niki Parmar, Jakob Uszkoreit, Llion Jones,
  Aidan~N Gomez, {\L}ukasz Kaiser, and Illia Polosukhin.
\newblock Attention is all you need.
\newblock In \emph{Advances in neural information processing systems}, pages
  5998--6008, 2017.

\bibitem[Vicol et~al.(2021)Vicol, Metz, and Sohl-Dickstein]{pmlr-v139-vicol21a}
Paul Vicol, Luke Metz, and Jascha Sohl-Dickstein.
\newblock Unbiased gradient estimation in unrolled computation graphs with
  persistent evolution strategies.
\newblock In Marina Meila and Tong Zhang, editors, \emph{Proceedings of the
  38th International Conference on Machine Learning}, volume 139 of
  \emph{Proceedings of Machine Learning Research}, pages 10553--10563. PMLR,
  18--24 Jul 2021.
\newblock URL \url{https://proceedings.mlr.press/v139/vicol21a.html}.

\bibitem[Wang et~al.(2017)Wang, Wu, Moore, and Russell]{wang2017meta}
Tongzhou Wang, Yi~Wu, David~A Moore, and Stuart~J Russell.
\newblock Meta-learning mcmc proposals.
\newblock \emph{arXiv preprint arXiv:1708.06040}, 2017.

\bibitem[Wichrowska et~al.(2017)Wichrowska, Maheswaranathan, Hoffman,
  Colmenarejo, Denil, de~Freitas, and Sohl-Dickstein]{wichrowska2017learned}
Olga Wichrowska, Niru Maheswaranathan, Matthew~W Hoffman, Sergio~Gomez
  Colmenarejo, Misha Denil, Nando de~Freitas, and Jascha Sohl-Dickstein.
\newblock Learned optimizers that scale and generalize.
\newblock \emph{International Conference on Machine Learning}, 2017.

\bibitem[Xiao et~al.(2017)Xiao, Rasul, and Vollgraf]{xiao2017/online}
Han Xiao, Kashif Rasul, and Roland Vollgraf.
\newblock Fashion-mnist: a novel image dataset for benchmarking machine
  learning algorithms, 2017.

\bibitem[Xiong and Hsieh(2020)]{xiong2020improved}
Yuanhao Xiong and Cho-Jui Hsieh.
\newblock Improved adversarial training via learned optimizer.
\newblock In \emph{European Conference on Computer Vision}, pages 85--100.
  Springer, 2020.

\bibitem[Xu et~al.(2017)Xu, Qin, Wang, and Liu]{xu2017reinforcement}
Chang Xu, Tao Qin, Gang Wang, and Tie-Yan Liu.
\newblock Reinforcement learning for learning rate control.
\newblock \emph{arXiv preprint arXiv:1705.11159}, 2017.

\bibitem[Xu et~al.(2019)Xu, Dai, Kemp, and Metz]{xu2019learning}
Zhen Xu, Andrew~M Dai, Jonas Kemp, and Luke Metz.
\newblock Learning an adaptive learning rate schedule.
\newblock \emph{arXiv preprint arXiv:1909.09712}, 2019.

\bibitem[You et~al.(2019)You, Li, Reddi, Hseu, Kumar, Bhojanapalli, Song,
  Demmel, Keutzer, and Hsieh]{you2019large}
Yang You, Jing Li, Sashank Reddi, Jonathan Hseu, Sanjiv Kumar, Srinadh
  Bhojanapalli, Xiaodan Song, James Demmel, Kurt Keutzer, and Cho-Jui Hsieh.
\newblock Large batch optimization for deep learning: Training bert in 76
  minutes.
\newblock \emph{arXiv preprint arXiv:1904.00962}, 2019.

\bibitem[Zeiler(2012)]{zeiler2012adadelta}
Matthew~D Zeiler.
\newblock Adadelta: an adaptive learning rate method.
\newblock \emph{arXiv preprint arXiv:1212.5701}, 2012.

\bibitem[Zheng et~al.(2022)Zheng, Chen, Hu, and Wang]{zheng2022symbolic}
Wenqing Zheng, Tianlong Chen, Ting-Kuei Hu, and Zhangyang Wang.
\newblock Symbolic learning to optimize: Towards interpretability and
  scalability.
\newblock In \emph{International Conference on Learning Representations
  (ICLR)}, 2022.

\end{thebibliography}
